\documentclass{article} 
\usepackage{iclr2027_arxiv,times}

\usepackage{amsmath,amsfonts,bm}

\def\eqref#1{equation~\ref{#1}}

\def\1{\bm{1}}

\DeclareMathAlphabet{\mathsfit}{\encodingdefault}{\sfdefault}{m}{sl}
\SetMathAlphabet{\mathsfit}{bold}{\encodingdefault}{\sfdefault}{bx}{n}

\DeclareMathOperator*{\argmin}{arg\,min}

\usepackage[utf8]{inputenc} 
\usepackage[T1]{fontenc}    
\usepackage{hyperref}       
\usepackage{url}            
\usepackage{booktabs}       
\usepackage{amsfonts}       
\usepackage{nicefrac}       
\usepackage{microtype}      
\usepackage{xcolor}         
\usepackage{colortbl}       
\usepackage{multirow}       
\usepackage{amsmath}
\usepackage{graphicx}
\usepackage{subcaption}

\usepackage{cleveref}

\usepackage[font=small,labelfont=bf]{caption}
\usepackage{tabularx}
\usepackage{adjustbox}

\newcommand{\benchscore}[2]{%
  \mbox{#1{\tiny\color{black!60}\,$\pm$#2}}%
}
\newcommand{\benchref}[1]{%
  \textcolor{black!60}{#1}{\tiny\phantom{\,$\pm$0.0}}%
}

\title{LRCC: Generalizing \textbf{L}ow-\textbf{R}ank Compression with \textbf{C}onditional \textbf{C}omputation}

\author{%
\makebox[\dimexpr\textwidth-2\tabcolsep\relax][c]{%
  Thomas V. Fontanari$^{*}$\hspace{3ex}Maximo Rulli\hspace{3ex}Federico Alvetreti}\\
\makebox[\dimexpr\textwidth-2\tabcolsep\relax][c]{%
  \bfseries Donatella Genovese\hspace{3ex}Simone Scardapane}\\
\makebox[\dimexpr\textwidth-2\tabcolsep\relax][c]{%
  \normalfont\fontsize{10}{12}\selectfont Sapienza University of Rome}%
}

\newcommand{\lrcc}{\textsc{LRCC}}
\newcommand{\lrccblk}{\textsc{LRCC}}
\newcommand{\swiftsvd}{\textsc{Swift-SVD}}
\newcommand{\llamaoneb}{Llama-3.2-1B}
\newcommand{\llamaSevenB}{Llama-2-7B}
\newcommand{\qwenThreeFourB}{Qwen3-4B}
\newcommand{\ie}{i.e.}

\newcommand{\wikitext}{WikiText-2}
\newcommand{\softcmidrule}{\arrayrulecolor{black!25}\cmidrule(lr){2-11}\arrayrulecolor{black}}
\newcommand{\err}[2]{%
  $#1_{\scriptscriptstyle \pm #2}$%
}

\newcommand{\NonWord}{\textsc{Non-Word}}

\iclrfinalcopy 
\begin{document}

\maketitle
\begingroup
\renewcommand{\thefootnote}{$*$}%
\footnotetext{Correspondence to thomas.fontanari@uniroma1.it}
\endgroup

\begin{abstract}
Low-rank compression reduces the cost of pretrained language models by replacing linear transformations with low-rank factorizations.
However, conventional methods use a fixed rank allocation during inference, assigning the same amount of compute regardless of the input token.
We introduce Low-Rank Conditional Computation (LRCC), which adds token-dependent computation to pretrained models by training one lightweight router per Transformer block to select among a small set of nested low-rank paths.
During training, the low-rank factors remain frozen, and only the routers are optimized. 
We evaluate \lrccblk{} on Llama and Qwen models for language modeling and zero-shot downstream tasks.
Within the same average active-parameter budget, \lrccblk{} improves the predictive performance over static low-rank compression, including a 7.6 percentage-point gain in average downstream accuracy on Llama-2-7B over static methods.
At matched batch-size-1 decoding latency, LRCC improves both perplexity and downstream accuracy on \llamaoneb{} and remains competitive on \llamaSevenB{}, without specialized kernels.
Finally, we assess the usefulness of assigning a token-wise path by analyzing the routers' path choices.
\end{abstract}

\section{Introduction}
Foundation models \citep{bommasani2021opportunities} achieve strong predictive performance across several tasks \citep{devlin2019bert, brown2020fsl, radford2021learning} but require substantial computational and memory resources, motivating efficiency techniques.
Low-rank compression methods \citep{chen2021drone} approximate the original weight matrices with low-rank products, typically obtained using SVD or its activation-aware variants.
Given activations collected on a calibration dataset, these methods seek a rank-constrained approximation that minimizes the reconstruction error of each layer's outputs.
Layer-specific ranks are then selected to satisfy a global budget via fixed heuristics or learned allocations \citep{qi2026swiftsvd,wang2025dobi}.
Conditional computation \citep{scardapane2024conditional,pfeiffer2023modular} instead adapts the computational graph to input activations. Applying it to pretrained models, however, can require additional training and compute modules alongside lightweight routers.

Previous work has demonstrated that SVD-like decompositions of linear layers can be used to produce many-in-one models, where the smaller ones correspond to a subset of the weights of larger ones \citep{yuan2025asvd, zaccone2026flexrank}.
The nested structure of these approaches provides an inexpensive basis for conditional computation.
Routing over this hierarchy at every linear layer would nevertheless require many routers, each choosing among thousands of possible ranks.
For example, a linear layer in \llamaSevenB{} can have up to $4096$ rank choices, and each Transformer block contains seven such layers.
The challenge is to expose useful flexibility without making routing itself expensive.

This work asks two questions: \emph{can token-dependent rank allocation improve predictive performance within the same active-parameter budget, and do these gains translate into better quality at matched decoding latency?}
We distinguish these questions because activating fewer parameters does not by itself guarantee lower latency.
We address these questions by proposing \textbf{L}ow-\textbf{R}ank \textbf{C}onditional \textbf{C}omputation (\lrccblk{}).
Through low-rank decomposition and rank allocation strategies, we construct a small set of models with shared weights and different computational costs.
The smaller models correspond to a subset of the ranks used in the larger ones, leading to a nested family of models.
A single router per Transformer block is then trained to select which subset of ranks to activate within the block, while the pretrained model and shared low-rank factors remain frozen.
By using a small number of submodels and limiting the routers to the block level, we are able to show that token-dependent computation can be recovered from an existing low-rank hierarchy by training only lightweight routers, while still producing meaningful quality gains under controlled compute budgets. 

Our main contributions are summarized below:
\begin{enumerate}
    \item We introduce \lrccblk{}, which combines shared low-rank factors with block-level routing to allocate ranks dynamically to each token.
    \item We show that \lrccblk{} improves predictive performance over static low-rank compression within the same active-parameter budget, with gains of up to $7.6$ percentage points in average downstream accuracy. It also achieves competitive quality at matched batch-size-1 decoding latency without specialized CUDA kernels.
    \item We introduce a target-cost annealing scheme that produces a family of router checkpoints in one progressive training run, and analyze how the number of paths, routing choices, and deployment costs affect the resulting models.
\end{enumerate}

\begin{figure}[t]
  \centering
  \includegraphics[width=\linewidth]{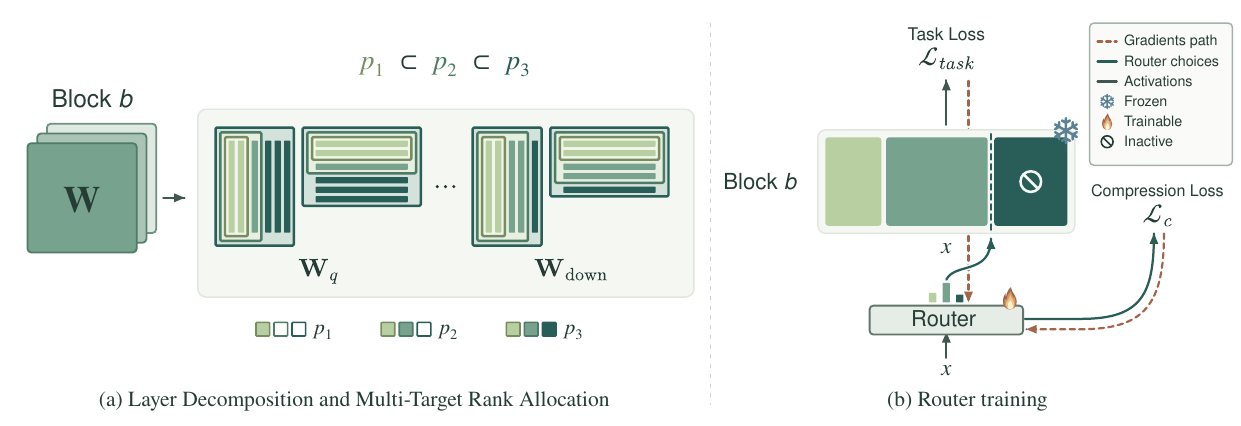}
  \caption{Overview of \lrccblk{}. First, (a) each linear weight matrix is decomposed into ordered rank-one components, and a rank-allocation algorithm constructs nested paths at different target costs. Then, (b) the low-rank factors are frozen and one router per block is trained to select a path for each token, balancing task and compression losses.}
  \label{fig:lrccBlkFramework}
\end{figure}

\section{Method}
Conventional low-rank compression methods \citep{yuan2025asvd, wang2025svd, parkina2026coala, qi2026swiftsvd, wang2025dobi} replace each linear weight matrix with a product of smaller matrices and retain only a fixed number of rank components in each layer.
The number retained in each layer is chosen so that the compressed model meets a predefined inference-cost budget, typically expressed as a fraction of the original model's cost.

In contrast, \lrccblk{} dynamically determines the number of ranks activated in each linear layer of a Transformer block for each token.
More precisely, we define multiple \textit{paths} through each Transformer block, where each path specifies the number of ranks activated in every linear layer of the block.
As in Matryoshka learning \citep{kusupati2022matryoshka}, the paths are ordered and nested: lower-cost paths retain subsets of the components used by higher-cost paths. 
Consequently, all paths share the same parameter base and can be selected using a single path index. 
We train only the block-level routers that choose among the possible paths for each token, while the remaining parameters are frozen. 
\Cref{fig:lrccBlkFramework} illustrates the framework.

\subsection{Layer Decomposition}\label{sec:layerDecomposition}
For each linear weight matrix $\mathbf{W}_{b\ell}\in\mathbb{R}^{d_{\mathrm{in}}\times d_{\mathrm{out}}}$ at layer $\ell$ in Transformer block $b$, we compute an activation-aware decomposition using a matrix of input activations $\mathbf{X}_{b\ell} \in \mathbb R^{n \times d_{\mathrm{in}}}$, where $n$ is the number of token activations collected on a calibration dataset. 
More formally, we solve for each layer
\begin{equation}
    \argmin_{\widetilde{\mathbf{W}}_{b\ell, r}}
    \left\|
        \mathbf{X}_{b\ell}\mathbf{W}_{b\ell}
        - \mathbf{X}_{b\ell}\widetilde{\mathbf{W}}_{b\ell, r}
    \right\|_F
    \quad
    \text{s.t.}
    \quad
    \operatorname{rank}(\widetilde{\mathbf{W}}_{b\ell,r}) \le r.
\label{eq:optimProblem}
\end{equation}
Let $\mathbf{V}_{b\ell} \in \mathbb R^{d_{\mathrm{out}} \times d_{\mathrm{out}}}$  denote the matrix of right singular vectors of $\mathbf{X}_{b\ell}\mathbf{W}_{b\ell}$ with columns ordered by decreasing singular value, and let $\mathbf{V}_{b\ell,:r}$ denote its first $r$ columns.
Then, \Cref{eq:optimProblem} has an analytical solution given by 
\begin{equation}
\widetilde{\mathbf{W}}_{b\ell,r} = \mathbf{W}_{b\ell}\mathbf{V}_{b\ell,:r} \mathbf{V}_{b\ell,:r}^{\top}.
\label{eq:solutionAASVD}
\end{equation}
Appendix~\ref{app:theory} proves optimality and nestedness and Appendix~\ref{app:efficientComputation} describes how to compute the factors efficiently.

\Cref{eq:solutionAASVD} has the important property that, similar to standard SVD, optimal solutions for different target ranks $r$ are nested.
Hence, we can obtain the best rank-$r$ activation-aware approximation by indexing the first $r$ columns of the precomputed matrices $\mathbf{A}_{b\ell} = \mathbf{W}_{b\ell} \mathbf{V}_{b\ell}$ and $\mathbf{B}_{b\ell,}=\mathbf{V}_{b\ell}$.
More concretely,
\begin{equation}
\widetilde{\mathbf{W}}_{b\ell,r}
=
(\mathbf{W}_{b\ell}\mathbf{V}_{b\ell,:r})
\mathbf{V}_{b\ell,:r}^{\top}
=
\mathbf{A}_{b\ell,:r}\mathbf{B}_{b\ell,:r}^{\top}.
\end{equation}
Naively replacing a weight matrix with the product of its low-rank factors does not necessarily reduce computation. 
A dense linear transformation requires approximately $2d_{\mathrm{in}}d_{\mathrm{out}}$ FLOPs per token, whereas its rank-$r$ factorization requires approximately $2r(d_{\mathrm{in}}+d_{\mathrm{out}})$ FLOPs, yielding a computational benefit only when
$r < \frac{d_{\mathrm{in}}d_{\mathrm{out}}}{d_{\mathrm{in}}+d_{\mathrm{out}}}$.

\subsection{Path Construction}\label{sec:pathConstruction}
In \Cref{sec:layerDecomposition}, we explained how to produce the decompositions of each linear transform in the model.
Crucially, the decompositions form a nested structure, where retaining more components cannot increase reconstruction error in \Cref{eq:optimProblem}, but induces a larger number of FLOPs.
Routing over every possible rank at every layer would introduce many routing decisions and a large choice set for each router.
Our path-count ablation in \Cref{sec:mainNumberPaths} shows that a small set of choices is sufficient to capture most of the quality gains.

Instead, we simplify the routing problem by placing routers only at the beginning of Transformer blocks and restricting them to a small set of choices. We first construct $T$ submodels at different target costs using the rank-allocation procedure of Swift-SVD \citep{qi2026swiftsvd}.
Given a target cost, Swift-SVD assigns a rank to each linear layer according to its estimated importance to model performance.

More formally, submodel $m$ with $m=1,...,T$ is defined by the rank allocation
\begin{equation*}
    \mathcal{R}_m = \{r_{b\ell m} \}_{b, \ell},
\end{equation*}
where $r_{b\ell m}$ denotes the rank retained by linear layer \(\ell\) in Transformer block \(b\) under submodel \(m\). 
Equivalently, each weight matrix of the submodel is approximated as
\begin{equation}
\widetilde{\mathbf{W}}_{b\ell}^{(m)}
=
\mathbf{A}_{b\ell,:r_{b\ell m}}\mathbf{B}_{b\ell,:r_{b\ell m}}^{\top}.
\end{equation}
Thus, each submodel specifies a different computational budget by assigning a particular rank to every linear layer.
These whole-model allocations define the choices available to each block-level router. 
In block \(b\), path \(p_{bm}\) is the restriction of submodel \(m\)'s allocation to that block,
$$ p_{bm} = \{r_{b\ell m}\}_{\ell}. $$
Selecting \(p_{bm}\) therefore determines simultaneously how many ranks are activated in each linear layer of block \(b\).

We can count the number of parameters in a path based on its set of ranks.
A factorized matrix of rank $r$ activates $r(d_{\mathrm{in}}+d_{\mathrm{out}})$ parameters, since it is represented by two factors of sizes $d_{\mathrm{in}}\times r$ and $d_{\mathrm{out}}\times r$.
Therefore, the parameter count of path $p_{bm}$ is
\begin{equation}
P_{bm}
=
\sum_{\ell}
r_{b\ell m}
\left(
d_{\mathrm{in},b\ell}+d_{\mathrm{out},b\ell}
\right).
\end{equation}

\lrccblk{} trains routers to assign each token to a corresponding path.
Specifically, for each token entering a Transformer block $b$, the router $g_b: \mathbb{R}^{d_h} \rightarrow [0,1]^{T}$ is a learnable linear transform that maps the input activations $h \in \mathbb{R}^{d_h}$ to a Gumbel-Softmax distribution over the $T$ available paths.
During both training and inference, a hard one-hot selection determines the executed path and hence the number of activated parameters.

\subsection{Router Training}\label{sec:mainRouterTraining}
The routers are optimized to minimize the task loss and a \textit{compression loss}, which we define in this section.
To preserve differentiability and retain the hard one-hot selection of the routers, we use the straight-through estimator (STE) \citep{jang2017categorical}.
During training, all pretrained model parameters and low-rank factors remain frozen, and only the routers are updated.
\lrccblk{} uses the ratio of activated parameters relative to the original model as the cost of processing a given token.
Let $c_b$ denote the average normalized cost of block $b$, and let $C_b$ denote the normalized cost of its original non-factorized implementation.
Because a low-rank factorization can activate more parameters than the original block, we softly cap its cost during training using a smooth approximate of $\min(c_b, C_b)$ given by 
\begin{equation}
    \bar{c}_b
    =
    c_b
    -
    \frac{C_b}{\kappa}
    \operatorname{softplus}\left(
        \frac{\kappa(c_b - C_b)}{C_b}
    \right),
\end{equation}
where $\kappa$ is a hyperparameter that determines the smoothness of the approximation.
This reflects that blocks whose factorized implementation is more expensive than the original one should be kept non-factorized when deployed, while avoiding the zero-gradient region introduced by a hard cap.

The global cost and corresponding compression loss are then
\begin{equation}
    c_{\mathrm{global}}
    =
    \sum_{b=1}^{B} \bar{c}_b,
    \qquad
    \mathcal{L}_c
    =
    \max\left(0, c_{\mathrm{global}} - t\right),
\end{equation}
where $t$ is the user-defined target cost.
The routers are then trained to jointly minimize the task loss $\mathcal{L}_{task}$ and the computational cost:
\begin{equation}
    \mathcal{L}
    =
    \mathcal{L}_{\mathrm{task}}
    +
    \lambda_s \mathcal{L}_{\mathrm{c}},
\end{equation}
where $\lambda_s$ is an adaptive weight updated at training step $s$ according to the current constraint violation, as described in \Cref{sec:annealing}.
Further details on the computation of $c_b$ and the soft cap are given in Appendix~\ref{app:routerTraining}.

\subsection{Target-cost Annealing}\label{sec:annealing}
Because the paths are nested, \lrccblk{} naturally supports producing routers for multiple target costs in a single progressive training run, moving from a high-cost model to one with fewer activated parameters.
We start with a high target cost, $0.95$ in our experiments, and train the router until convergence (or until a maximum number of epochs is reached).
Then, we reduce the target cost to a slightly lower value, $0.925$ in our case, and continue training the router weights to produce a new checkpoint.
We stop when the final target cost is reached, or when convergence fails for some target, i.e., the training process failed to obtain a router inducing a cost within an acceptable distance of the current target.

As compression becomes more aggressive, the task's loss increases and can dominate the total cost, making it difficult to reach a checkpoint sufficiently close to the target.
To avoid this, we introduce an adaptive compression-loss weight $\lambda_s$, which is updated after each optimization step based on the constraint violation:
\begin{equation}
    \lambda_{s+1} = \max(0, \lambda_s + \alpha (c_{\mathsf{global}} - t)),
\end{equation}
where $\alpha$ is a hyperparameter. 
Intuitively, when \(c_{\mathsf{global}} > t\), the compression-loss weight increases, and when \(c_{\mathsf{global}} < t\), it decreases, allowing the router to activate more parameters.

\section{Results}\label{sec:results}
\textbf{Models.}
We evaluate the performance of \lrccblk{} on \llamaoneb{} \citep{meta2024llama32}, \llamaSevenB{} \citep{touvron2023llama2} and \qwenThreeFourB{} \citep{yang2025qwen3} models, as they allow us to compare with the low-rank baseline methods \citep{qi2026swiftsvd, parkina2026coala, wang2025svd}.

\textbf{LRCC setup.}
We decompose all linear layers in each Transformer block, including both attention and MLP projections, as described in \Cref{sec:layerDecomposition}, and compute the ranks assigned to each block path using a rank-allocation algorithm, as described in \Cref{sec:pathConstruction}.
In all cases, we follow previous work and use 256 2,048-token samples from the training set of \wikitext{} \citep{merity2016pointer} to decompose the weights and train the router. 
We use the standard causal language-modeling loss as the task loss $\mathcal{L}_{\mathrm{task}}$.
Training starts at a target cost of 0.975, and we progressively obtain checkpoints down to 0.3 in steps of 0.025 using the annealing scheme described in \Cref{sec:annealing}.
Finally, during evaluation, we use 64 \wikitext{} validation samples to estimate the average activated parameter count of each block.
Blocks for which this exceeds the parameter count of the corresponding block with non-factorized linear layers are deployed in non-factorized form.

\textbf{Datasets.} 
The resulting models are evaluated on the \wikitext{} test set for language modeling and on several zero-shot downstream performance using ARC-Easy and ARC-Challenge \citep{clark2018arc}, BoolQ \citep{clark2019boolq}, HellaSwag \citep{zellers2019hellaswag}, OpenBookQA \citep{mihaylov2018openbookqa}, PIQA \citep{bisk2020piqa}, and WinoGrande \citep{sakaguchi2020winogrande}.

\textbf{Static baselines.} 
We compare \lrccblk{} with several state-of-the-art low-rank compression methods, namely SVD-LLM \citep{wang2025svd}, COALA \citep{parkina2026coala} and Swift-SVD \citep{qi2026swiftsvd}.
Since COALA and SVD-LLM do not propose a rank allocation method, we also evaluate variants using ranks obtained with the \swiftsvd{} rank allocation strategy.
We omit SVD-LLM's optional fine-tuning stage to isolate the effect of the compression method under a matched adaptation budget, since the other baselines do not use post-compression fine-tuning of the weights and an analogous refinement stage could likewise be applied to \lrccblk{} and the baselines.
See Appendix \ref{app:additionalResults} for details and additional results.


\begin{table*}[!t]
\centering
\begingroup
\normalfont
\footnotesize
\fontfamily{ptm}\selectfont
\setlength{\tabcolsep}{1.8pt}
\renewcommand{\arraystretch}{1.14}
\begin{adjustbox}{max width=\linewidth}
\begin{tabular}{l l<{\hspace{3pt}} *{8}{>{\scriptsize}r} r}
\toprule
& & \multicolumn{1}{c}{PPL $\downarrow$} & \multicolumn{8}{c}{Commonsense reasoning accuracy (\%) $\uparrow$} \\
\cmidrule(lr){3-3}\cmidrule(l){4-11}
Ratio & Method & \multicolumn{1}{c}{Wiki2} & \multicolumn{1}{c}{ARC-C} & \multicolumn{1}{c}{ARC-E} & \multicolumn{1}{c}{BoolQ} & \multicolumn{1}{c}{Hella} & \multicolumn{1}{c}{OBQA} & \multicolumn{1}{c}{PIQA} & \multicolumn{1}{c}{Wino} & \multicolumn{1}{c}{\textbf{Avg.}} \\
\midrule
\textcolor{black!60}{100\%} & \textcolor{black!60}{Base} & \textcolor{black!60}{5.49} & \benchref{44.80} & \benchref{73.95} & \benchref{79.42} & \benchref{76.08} & \benchref{44.00} & \benchref{78.62} & \benchref{69.30} & \benchref{66.60} \\
\midrule
\multirow{7}{*}[-1pt]{90\%} & COALA & 7.27 & \benchscore{35.5}{1.4} & \benchscore{57.5}{1.0} & \benchscore{49.3}{0.9} & \benchscore{62.8}{0.5} & \benchscore{38.4}{2.2} & \benchscore{70.6}{1.1} & \benchscore{65.2}{1.3} & \benchscore{54.2}{0.5} \\
 & SVD-LLM & 7.27 & \benchscore{36.2}{1.4} & \benchscore{57.4}{1.0} & \benchscore{50.1}{0.9} & \benchscore{62.8}{0.5} & \benchscore{39.0}{2.2} & \benchscore{70.8}{1.1} & \benchscore{65.3}{1.3} & \benchscore{54.5}{0.5} \\
 & SVD-LLM+Swift & 7.33 & \benchscore{35.5}{1.4} & \benchscore{58.3}{1.0} & \benchscore{55.0}{0.9} & \benchscore{62.8}{0.5} & \benchscore{38.4}{2.2} & \benchscore{71.1}{1.1} & \benchscore{62.9}{1.4} & \benchscore{54.8}{0.5} \\
 & COALA+Swift & 7.27 & \benchscore{35.1}{1.4} & \benchscore{58.4}{1.0} & \benchscore{57.4}{0.9} & \benchscore{62.8}{0.5} & \benchscore{38.4}{2.2} & \benchscore{70.9}{1.1} & \benchscore{63.1}{1.4} & \benchscore{55.1}{0.5} \\
 & Swift-SVD & 7.33 & \benchscore{35.6}{1.4} & \benchscore{58.7}{1.0} & \benchscore{55.4}{0.9} & \benchscore{62.8}{0.5} & \benchscore{39.6}{2.2} & \benchscore{70.7}{1.1} & \benchscore{62.9}{1.4} & \benchscore{55.1}{0.5} \\
\addlinespace[2pt]
\rowcolor{black!5}
\cellcolor{white} & \textbf{\lrccblk{}-5} & 6.17 & \benchscore{\textbf{41.8}}{1.4} & \benchscore{68.5}{1.0} & \benchscore{74.4}{0.8} & \benchscore{71.0}{0.5} & \benchscore{\textbf{41.4}}{2.2} & \benchscore{\textbf{74.8}}{1.0} & \benchscore{66.3}{1.3} & \benchscore{62.6}{0.5} \\
\rowcolor{black!5}
\cellcolor{white} & \textbf{\lrccblk{}-9} & \textbf{6.03} & \benchscore{40.4}{1.4} & \benchscore{\textbf{68.6}}{1.0} & \benchscore{\textbf{74.9}}{0.8} & \benchscore{\textbf{71.7}}{0.4} & \benchscore{40.8}{2.2} & \benchscore{\textbf{74.8}}{1.0} & \benchscore{\textbf{67.8}}{1.3} & \benchscore{\textbf{62.7}}{0.5} \\
\midrule
\multirow{7}{*}[-1pt]{70\%} & COALA & 10.75 & \benchscore{28.0}{1.3} & \benchscore{43.9}{1.0} & \benchscore{41.2}{0.9} & \benchscore{46.1}{0.5} & \benchscore{\textbf{35.0}}{2.1} & \benchscore{62.1}{1.1} & \benchscore{59.3}{1.4} & \benchscore{45.1}{0.5} \\
 & SVD-LLM & 10.75 & \benchscore{27.2}{1.3} & \benchscore{44.2}{1.0} & \benchscore{41.6}{0.9} & \benchscore{46.1}{0.5} & \benchscore{34.6}{2.1} & \benchscore{61.8}{1.1} & \benchscore{59.2}{1.4} & \benchscore{45.0}{0.5} \\
 & SVD-LLM+Swift & 10.58 & \benchscore{27.7}{1.3} & \benchscore{44.1}{1.0} & \benchscore{41.9}{0.9} & \benchscore{46.6}{0.5} & \benchscore{34.8}{2.1} & \benchscore{61.9}{1.1} & \benchscore{60.5}{1.4} & \benchscore{45.4}{0.5} \\
 & COALA+Swift & 10.58 & \benchscore{27.5}{1.3} & \benchscore{44.1}{1.0} & \benchscore{42.0}{0.9} & \benchscore{46.4}{0.5} & \benchscore{34.4}{2.1} & \benchscore{61.6}{1.1} & \benchscore{60.3}{1.4} & \benchscore{45.2}{0.5} \\
 & Swift-SVD & 10.58 & \benchscore{28.1}{1.3} & \benchscore{44.2}{1.0} & \benchscore{43.1}{0.9} & \benchscore{46.5}{0.5} & \benchscore{\textbf{35.0}}{2.1} & \benchscore{61.1}{1.1} & \benchscore{59.4}{1.4} & \benchscore{45.4}{0.5} \\
\addlinespace[2pt]
\rowcolor{black!5}
\cellcolor{white} & \textbf{\lrccblk{}-5} & 9.49 & \benchscore{31.4}{1.4} & \benchscore{46.3}{1.0} & \benchscore{\textbf{46.6}}{0.9} & \benchscore{56.0}{0.5} & \benchscore{34.6}{2.1} & \benchscore{\textbf{67.8}}{1.1} & \benchscore{\textbf{61.6}}{1.4} & \benchscore{\textbf{49.2}}{0.5} \\
\rowcolor{black!5}
\cellcolor{white} & \textbf{\lrccblk{}-9} & \textbf{9.05} & \benchscore{\textbf{31.9}}{1.4} & \benchscore{\textbf{46.4}}{1.0} & \benchscore{41.7}{0.9} & \benchscore{\textbf{56.5}}{0.5} & \benchscore{34.4}{2.1} & \benchscore{65.1}{1.1} & \benchscore{61.5}{1.4} & \benchscore{48.2}{0.5} \\
\bottomrule
\end{tabular}%
\end{adjustbox}
\endgroup
\caption{Predictive performance of \llamaSevenB{} within the indicated active-parameter budgets. LRCC achieves the highest average accuracy at all compression ratios. Results for Llama-2-7B compressed using 256 samples of sequence length 2048.
Avg. reports mean accuracy over the seven reasoning tasks; Wiki2 reports perplexity.}
\label{tab:model-benchmarks-llama27b-256-seqlen2048}
\end{table*}

\subsection{Predictive Performances}\label{sec:lm-performance}
\Cref{tab:model-benchmarks-llama27b-256-seqlen2048} shows the predictive performance results across selected ratios for \llamaSevenB{} (see Appendix~\ref{app:extraPerformance} for results on \llamaoneb{} and \qwenThreeFourB{}).
Since \lrccblk{} is dynamic, its actual average ratio of activated parameters slightly differs from the specified target costs. 
In cases where the actual cost was higher on a task than the target cost, we show the performance obtained using a router with lower target costs, to ensure a fairer comparison.
The actual average costs for each model in each task shown are specified in \Cref{tab:deploy-cost-models-lrcc5-9-256-seqlen2048}.
In \Cref{sec:mainAvgActParams}, we show that the actual cost of the \lrccblk{} models follow closely their target costs across several datasets not seen during training.
At the $90\%$ budget, \lrccblk{}-9 reaches $62.7\%$ average accuracy, compared with $55.1\%$ for the best static baseline, a gain of $7.6$ percentage points.
At the $70\%$ budget, \lrccblk{}-5 improves average accuracy from $45.4\%$ to $49.2\%$.
Both variants also improve \wikitext{} perplexity at these budgets.
The gains extend across most individual tasks, showcasing the beneficial effect of introducing conditional computation to the low-rank decompositions.
Several static baselines obtain similar results among themselves, particularly on \llamaSevenB{}.
This is consistent with their shared activation-aware reconstruction objective: in exact arithmetic and under the same calibration and rank allocation, their decompositions target equivalent optimal output approximations, while their numerical procedures differ.
Among the evaluated combinations, COALA+Swift generally gives the strongest static results on Llama, so we use it for the trade-off curves.

\begin{figure}[h]
  \centering
  \includegraphics[width=\linewidth]{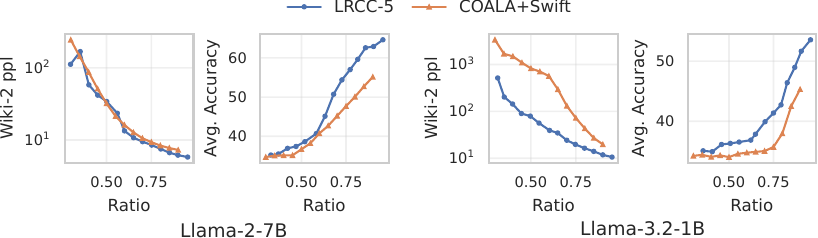}
  \caption{Predictive performance versus actual active-parameter ratio for \lrccblk{}-5 and COALA+Swift. Average accuracy is the mean over the seven downstream tasks.}
  \label{fig:paretoActiveRatio}
\end{figure}

\Cref{fig:paretoActiveRatio} compares \lrccblk{}-5 with COALA+Swift over a wider range of measured active-parameter ratios.
We evaluate checkpoints with target costs from $0.3$ to $0.95$ in steps of $0.05$ and plot their actual average costs on each dataset.
The path-count ablation in \Cref{sec:mainNumberPaths} motivates using five paths and Appendix~\ref{sec:appendixNumberPaths} gives additional results.
\lrccblk{} improves the quality--cost trade-off in almost all plotted settings, with particularly large gains on the downstream tasks.
We hypothesize that aggressive static truncation removes subspaces important for particular tasks, leading to sharp performance drops \citep{ahmad2025beyond}.
\lrccblk{} can retain such components for selected tokens while meeting an average budget, providing a possible explanation for the gains.
We also observe a greater language-modeling benefit on \llamaoneb{}.
One possible explanation is that larger models are more robust to compression \citep{yuan2025asvd}.

To test whether the quality gains extend beyond Llama, we compare \lrccblk{} with \swiftsvd{} on \qwenThreeFourB{}.
Appendix~\ref{app:extraPerformance} reports the results and the measured-cost curves in \Cref{fig:paretoActiveRatioQwen}.
\lrccblk{} improves perplexity and average downstream accuracy across the reported budgets.

\subsection{Computational Performance}\label{sec:mainComputePerformance}
\begin{figure}[h]
  \centering
  \includegraphics[width=\linewidth]{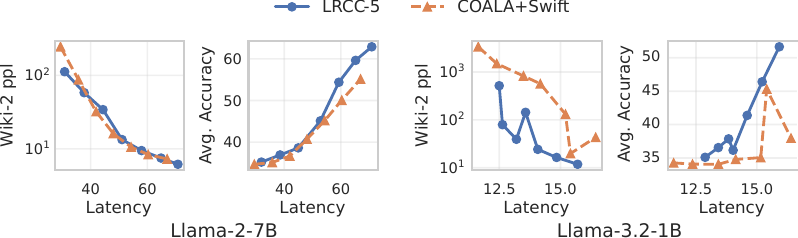}
  \caption{Pareto curves of \lrccblk{}-5 and COALA+Swift across various decoding latencies (ms/token) applied to \llamaSevenB{} and \llamaoneb{}. The average accuracy is obtained by averaging the results on the commonsense reasoning downstream tasks.}
  \label{fig:paretoLatency}
\end{figure}

We also measured the latency for models with target costs from \(0.3\) to \(0.9\) on samples from the test sets. 
Both LRCC and the baseline were compiled with TorchDynamo and TorchInductor using identical settings. 
Following \citet{lin2024awq}, we measured batch-size-1 autoregressive decoding, which is a setting relevant to local, resource-constrained deployment.
Each measurement was repeated 5 times on a DGX Spark (NVIDIA GB10 Grace Blackwell) with 128\,GB of unified memory.
We teacher-forced the ground-truth task tokens through the models to ensure identical inputs for both deployments and measured latency as the time required to decode each complete sequence divided by its number of decoded tokens. 
For multiple-choice tasks, we timed each candidate continuation and treated all continuations associated with the same document as one sample cluster.
\Cref{fig:paretoLatency} shows the resulting latency--quality trade-offs.
Appendix~\ref{sec:addResCompute} gives the latency tables and measurement protocol; Appendix~\ref{sec:appendixNumberPaths} reports compilation and prefill times.
\lrccblk-5 achieves better predictive performance at the same latency on all tasks for \llamaoneb{} and on downstream tasks for \llamaSevenB{} at higher latencies. 
The results are mixed for \llamaSevenB{} on \wikitext{}.

\section{Analysis}
\lrccblk{} assigns variable amounts of compute to different input samples by choosing a different path among available options.
We now ask how many paths are needed, whether the routers use them differently across tokens, and whether the learned budgets transfer beyond the router-training data.

\subsection{Effect of the Number of Paths}\label{sec:mainNumberPaths}
\begin{figure}[h]
  \centering
  \begin{subfigure}[t]{0.44\textwidth}
    \centering
    \includegraphics[width=\linewidth]{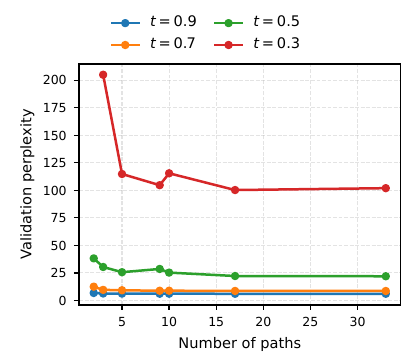}
    \caption{}
  \end{subfigure}
  \hfill
  \begin{subfigure}[t]{0.44\textwidth}
    \centering
    \includegraphics[width=\linewidth]{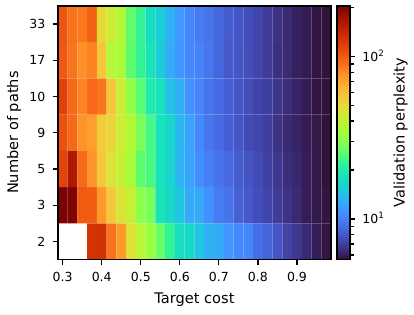}
    \caption{}
  \end{subfigure}
  \caption{
    Ablation on the number of paths for \llamaSevenB. (a) shows \wikitext{} validation perplexity as a function of the number of paths for selected target costs, while (b) shows the results for all target costs from $0.3$ to $0.975$ in steps of $0.025$. Hyperparameter configurations for which no run converged are shown in white.
  }
  \label{fig:numberPathsLlama2}
\end{figure}

\lrccblk{} requires determining the number of available paths within each block.
To better understand the influence of this hyperparameter, we trained models with \(T\in\{2,3,5,9,10,17,33\}\) paths per block and measured perplexity on the \wikitext{} validation set.
\Cref{fig:numberPathsLlama2} shows the validation perplexity obtained across several target costs and number of paths.
Perplexity increases sharply when only two paths are available and remains notably higher with three paths.
The effects are particularly pronounced at lower target costs, suggesting that the limited granularity in these settings makes it harder to find a feasible solution.
Increasing the number of paths beyond five provides comparatively little improvement in most settings.
Additional results for \llamaoneb{}, and Torch compilation, prefill and decoding times across various path cardinalities are shown in Appendix~\ref{sec:appendixNumberPaths}.

\subsection{Token-dependent Routing}\label{sec:routerChoices}
\begin{figure}[h]
  \centering
  \begin{subfigure}[t]{0.470\textwidth}
    \centering
    \includegraphics[width=\linewidth]{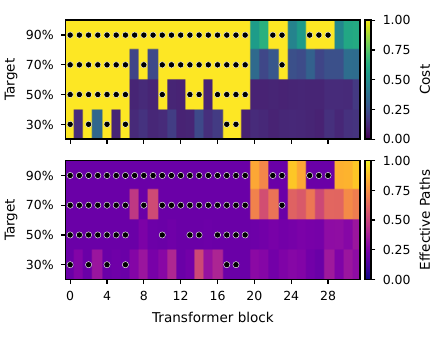}
    \caption{}
  \end{subfigure}
  \begin{subfigure}[t]{0.508\textwidth}
    \centering
    \includegraphics[width=\linewidth]{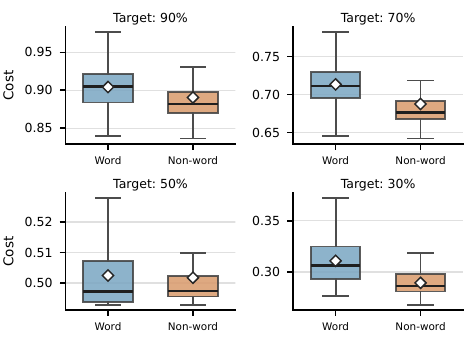}
    \caption{}
  \end{subfigure}
  \caption{Router choices for \llamaSevenB{} on \wikitext{}. (a) Average per-block cost (top) and effective fraction of paths (bottom) for each target. The dots indicate blocks that are deployed without factorization. (b) Distributions of the costs assigned to word and non-word tokens for each target.}
  \label{fig:routerChoicesLlama2}
\end{figure}

To test whether the routers collapse to a single path for all tokens, we analyzed their choices for 64 samples from \wikitext{}, each comprising 2,048 tokens.
Specifically, we measured the average computational cost across each block and the effective fraction of paths at different target costs.
For each router, we define the effective fraction of paths as
\begin{equation}
    F_{\mathrm{eff}} = \frac{\exp(H(f))}{T},
\end{equation}
where \(f\) denotes the empirical distribution of path selections across tokens, \(H(f)\) is its entropy computed with natural logarithms, and \(T\) is the number of available paths.
This quantity equals \(1\) when all paths are selected with equal frequency and \(1/T\) when all tokens are assigned to a single path.
\Cref{fig:routerChoicesLlama2}(a) shows the results for \llamaSevenB{}.
The results show a backbone of blocks retained in dense form, alongside blocks that route tokens through different low-rank paths.
As the target cost decreases, the number of blocks that exhibit conditional routing increases.
Multiple paths remain in use, with some Transformer blocks using almost all available paths.
Together, these observations show that the routed blocks do not collectively collapse to a static rank allocation, even without an explicit path-balancing loss.

We also analyze whether certain tokens were more often assigned to low-cost paths.
As shown in \Cref{fig:routerChoicesLlama2}(b), non-word tokens (i.e., tokens containing no alphanumeric characters) are more frequently assigned to lower-cost paths.
This provides evidence that compute allocation varies with token type. 
Appendix~\ref{app:routerChoices} reports similar patterns for \llamaoneb{}.

\subsection{Target vs. Actual Deployment Cost}\label{sec:mainAvgActParams}
\begin{figure*}[!ht]
  \centering
  \includegraphics[width=0.9\textwidth]{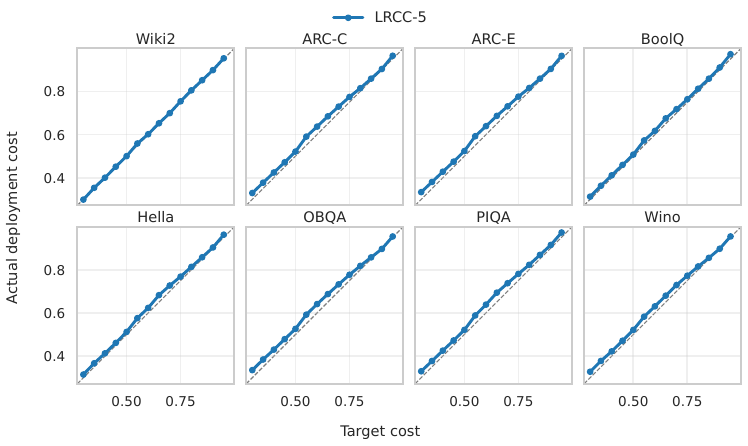}
  \caption{Target versus actual deployment cost for \llamaSevenB{} with \lrccblk{}-5. \lrccblk{} costs remain close to the target across different datasets.}
  \label{fig:deploy-cost-target-vs-actual-llama27b}
\end{figure*}

\lrccblk{} routers are fitted to use a certain amount of compute on average, but this is not strictly enforced during inference.
Hence, we measured the actual ratio of activated parameters of \lrccblk{} across different datasets used for testing.
\Cref{fig:deploy-cost-target-vs-actual-llama27b} shows the results for \llamaSevenB{}.
Although the routers were trained only on \wikitext{}, actual deployment costs remain close to their targets across the evaluated datasets.
Appendix~\ref{sec:avgActParams} presents the results for \llamaoneb{} and \qwenThreeFourB{}.

\section{Related Work}
\paragraph{Low-rank Compression} Several low-rank compression methods have been proposed in recent years \citep{shihab-etal-2026-beyond,qi2026swiftsvd,parkina2026coala,wang2025svd,yuan2025asvd,wang2025dobi}.
\citet{zaccone2026flexrank} and \citet{rauba2026deep} construct nested low-rank submodels to support multiple compute budgets within a single model.
Concurrent work by \citet{zhu2026different} selects prompt-dependent singular components under fixed per-layer rank budgets and reuses the selected subspaces across decoding steps.
These approaches offer either selectable global budgets or prompt-dependent subspaces. \lrccblk{} instead varies the active rank at each block for each token, under an average cost target.

\paragraph{Conditional Computation} Mixture-of-Experts \citep{shazeer2017} has become standard in LLM pretraining, and several works have proposed architectures with dynamic computational graphs (e.g., \citep{raposo2024mixture, ainslie2023colt5, devvrit2024matformer}).
However, these approaches generally require training the model from scratch.
Like \lrccblk{}, some approaches adapt pretrained checkpoints.
Moefication and upcycling are arguably among the best known \citep{zhang2022moefication, komatsuzaki2023sparse, szatkowski2024exploiting}, but other forms have been studied (e.g. \citep{wojcik2025adaptive, cai2024flextron}).
These techniques are generally either difficult to apply to existing models (e.g., Moefication is tailored to ReLU-induced activation sparsity) or require training additional compute modules, whereas \lrccblk{} trains only lightweight routers over fixed factors.

\section{Conclusion}\label{sec:conclusion}
In this work, we introduced \lrccblk{}, a technique for adding conditional computation to pretrained Transformers by training block-level routers over shared low-rank factors.
\lrccblk{} improves predictive performance within fixed active-parameter budgets on Llama models, with gains of up to $7.6$ percentage points in average downstream accuracy; the Qwen experiments extend the quality gains to another model family.
At matched batch-size-1 decoding latency, \lrccblk{} improves perplexity and average downstream accuracy on \llamaoneb{} and remains competitive on \llamaSevenB{}, without custom CUDA kernels.
Our analysis shows that a small set of paths captures most of the quality gains and that the routers retain token-dependent path choices without an explicit balancing loss.
Together, these results establish shared low-rank hierarchies as a practical basis for introducing conditional computation into pretrained models.

\subsection*{AI use statement}
We used generative AI tools, including coding agents, to discuss, critique, and refine our methodological ideas, improve the exposition of proof sketches, and assist with implementing code, given our specifications of the algorithms and requirements.
We also used generative AI tools for literature search and to revise text passages for clarity and presentation.
AI tools were not used to independently formulate the proposed method, design or run experiments, interpret experimental results, or generate experimental claims.
AI-assisted code was manually inspected, mathematical arguments were independently verified, and citations were checked against the original sources.
We take responsibility for the final content of the paper and all AI-assisted work.

\subsection*{Ethics statement}
This paper presents work whose goal is to advance the field of Machine Learning. There are many potential societal consequences of our work, none of which we feel must be specifically highlighted here.

\subsection*{Reproducibility statement}
\Cref{sec:layerDecomposition} describes the activation-aware low-rank decomposition, \Cref{sec:pathConstruction} the construction of block-level paths, \Cref{sec:mainRouterTraining} the router training objective, and \Cref{sec:annealing} the target-cost annealing procedure.
Appendix~\ref{app:theory} proves the low-rank approximation result.
Appendix~\ref{app:routerTraining} describes the cost computation and compression objective for training, and Appendix~\ref{app:implementationDetails} reports implementation details, training settings, and baseline hyperparameters.
Appendix~\ref{sec:addResCompute} describes the latency-measurement protocol.


\subsubsection*{Acknowledgments}
We acknowledge ISCRA for awarding this project access to the LEONARDO supercomputer, owned by the EuroHPC Joint Undertaking, hosted by CINECA (Italy).
Donatella Genovese is supported by PNRR MUR project PE0000013-FAIR.

\bibliography{iclr2027_conference}
\bibliographystyle{iclr2027_conference}

\newpage

\appendix
\section{Limitations}\label{app:limitations}
Our work focuses on demonstrating that conditioning the activated ranks on input activations can improve the quality--compute trade-off of low-rank compressed models, but several directions remain open.
First, we have not evaluated \lrccblk{} on nested subspace models that refine the factorized weights through fine-tuning, such as FlexRank \citep{zaccone2026flexrank}.
Combining such refinement with conditional rank selection may yield further gains.
Second, we demonstrated latency improvements only for batch-size-1 decoding, and
future work should investigate optimized implementations for prefill and larger batch
sizes.
Finally, our paths are constructed before router training from a fixed set of nested
rank allocations.
Jointly learning the path structure and the routing policy, or exploring more flexible
routing granularities, could further improve the allocation of computation across
inputs.

\section{Low-rank Approximation Results}\label{app:theory}
This section establishes the optimality and nestedness of the activation-aware rank-$r$ approximation used in \Cref{sec:layerDecomposition}, and its connection to reduced-rank regression \citep{izenman1975reduced}. Efficient computation of the factors is discussed in Appendix~\ref{app:efficientComputation}.

\paragraph{Optimality and nestedness.}
For simplicity, we omit the block and layer subscripts.
Let $\mathbf{W}\in\mathbb{R}^{d_{\mathrm{in}}\times d_{\mathrm{out}}}$ be a
linear weight matrix and let
$\mathbf{X}\in\mathbb{R}^{n\times d_{\mathrm{in}}}$ contain its calibration
inputs. We consider the rank-constrained reconstruction problem
\begin{equation}
    \min_{\widetilde{\mathbf{W}}_r}
    \left\|
        \mathbf{X}\mathbf{W}
        -
        \mathbf{X}\widetilde{\mathbf{W}}_r
    \right\|_F
    \qquad
    \text{s.t.}\quad
    \operatorname{rank}(\widetilde{\mathbf{W}}_r)\leq r.
    \label{eq:low_rank_problem_appendix}
\end{equation}
Let
\begin{equation}
    \mathbf{Y}
    =
    \mathbf{X}\mathbf{W}
    =
    \mathbf{U}\mathbf{\Sigma}\mathbf{V}^{\top}
\end{equation}
be an SVD of the layer outputs, with singular values ordered in decreasing order, and let $\mathbf{V}_{:r}$ denote the first $r$
columns of $\mathbf{V}$. We define
\begin{equation}
    \widetilde{\mathbf{W}}_r
    =
    \mathbf{W}\mathbf{V}_{:r}\mathbf{V}_{:r}^{\top}.
    \label{eq:optimal_weight_approx}
\end{equation}
Since
$\operatorname{rank}(\mathbf{W}\mathbf{V}_{:r}
\mathbf{V}_{:r}^{\top})\leq r$,
this is a feasible solution to
Equation~\ref{eq:low_rank_problem_appendix}. Its output on the
calibration inputs is
\begin{align}
    \mathbf{X}\widetilde{\mathbf{W}}_r
    &=
    \mathbf{X}\mathbf{W}
    \mathbf{V}_{:r}\mathbf{V}_{:r}^{\top} \\
    &=
    \mathbf{Y}\mathbf{V}_{:r}\mathbf{V}_{:r}^{\top} \\
    &=
    \mathbf{U}_{:r}\mathbf{\Sigma}_{:r}
    \mathbf{V}_{:r}^{\top}
    = \mathbf{Y}_r,
\end{align}
where $\mathbf{Y}_r$ is the truncated rank-$r$ SVD of
$\mathbf{Y}$. For any other feasible matrix
$\widehat{\mathbf{W}}_r$,
\begin{equation}
    \operatorname{rank}(\mathbf{X}\widehat{\mathbf{W}}_r)
    \leq
    \operatorname{rank}(\widehat{\mathbf{W}}_r)
    \leq r.
\end{equation}
Therefore, by the Eckart--Young--Mirsky theorem,
\begin{equation}
    \left\|
        \mathbf{X}\mathbf{W}
        -
        \mathbf{X}\widetilde{\mathbf{W}}_r
    \right\|_F
    =
    \left\|
        \mathbf{Y}-\mathbf{Y}_r
    \right\|_F
    \leq
    \left\|
        \mathbf{X}\mathbf{W}
        -
        \mathbf{X}\widehat{\mathbf{W}}_r
    \right\|_F.
\end{equation}
Hence, $\widetilde{\mathbf{W}}_r$ is a global optimum of
Equation~\ref{eq:low_rank_problem_appendix}.

Moreover, define the full factors
\begin{equation}
    \mathbf{A} = \mathbf{W}\mathbf{V},
    \qquad
    \mathbf{B} = \mathbf{V}.
\end{equation}
For every target rank $r$, the optimal approximation can then be
obtained by indexing the first $r$ columns of these same factors:
\begin{equation}
    \widetilde{\mathbf{W}}_r
    =
    \mathbf{A}_{:r}\mathbf{B}_{:r}^{\top}
    =
    (\mathbf{W}\mathbf{V}_{:r})
    \mathbf{V}_{:r}^{\top}.
    \label{eq:nested_optimal_factors}
\end{equation}
Thus, a single pair of ordered factors $\mathbf{A}$ and $\mathbf{B}$
simultaneously contains a globally optimal solution for every target
rank. In particular, the solutions are nested, since increasing the
rank from $r$ to $r+1$ only adds the next rank-one component,
\begin{equation}
    \widetilde{\mathbf{W}}_{r+1}
    =
    \widetilde{\mathbf{W}}_r
    +
    (\mathbf{W}\mathbf{v}_{r+1})\mathbf{v}_{r+1}^{\top}.
\end{equation}

\paragraph{Connection to reduced-rank regression.}
Equation~\ref{eq:low_rank_problem_appendix} can also be viewed as a
special instance of reduced-rank regression (RRR)
\citep{izenman1975reduced}. Given predictors $\mathbf{X}$ and
multivariate responses $\mathbf{Y}$, RRR solves
\begin{equation}
    \min_{\mathbf{C}}
    \|\mathbf{Y}-\mathbf{X}\mathbf{C}\|_F
    \qquad
    \text{s.t.}\quad
    \operatorname{rank}(\mathbf{C})\leq r.
\end{equation}
Let $\mathbf{C}_{\mathrm{OLS}}$ be an ordinary least-squares solution
and let
$\widehat{\mathbf{Y}}_{\mathrm{OLS}}
=\mathbf{X}\mathbf{C}_{\mathrm{OLS}}$.
If $\mathbf{V}_{:r}$ contains the leading $r$ right singular vectors
of $\widehat{\mathbf{Y}}_{\mathrm{OLS}}$, a reduced-rank solution is
\begin{equation}
    \mathbf{C}_{r}
    =
    \mathbf{C}_{\mathrm{OLS}}
    \mathbf{V}_{:r}\mathbf{V}_{:r}^{\top}.
\end{equation}
In our setting the responses are exactly the outputs of the known
pretrained linear transform,
$\mathbf{Y}=\mathbf{X}\mathbf{W}$.
Consequently, $\mathbf{W}$ itself is an ordinary least-squares
solution with zero residual, so that
$\widehat{\mathbf{Y}}_{\mathrm{OLS}}=\mathbf{X}\mathbf{W}$.
Substituting
$\mathbf{C}_{\mathrm{OLS}}=\mathbf{W}$ into the RRR solution therefore
recovers
\begin{equation}
    \mathbf{C}_{r}
    =
    \mathbf{W}\mathbf{V}_{:r}\mathbf{V}_{:r}^{\top}
    =
    \widetilde{\mathbf{W}}_r,
\end{equation}
providing an alternative interpretation of
Equation~\ref{eq:optimal_weight_approx}.

\section{Router Training Details}\label{app:routerTraining}
\subsection{Router Training}
For each token representation entering block $b$, the router produces a probability distribution over its $T$ paths. 
During both training and inference, the forward pass selects a single path using the highest router probability.
The executed path and its parameter count are computed from this hard one-hot selection, while the straight-through estimator (STE) propagates gradients through the soft probabilities during training.

To encourage the router to meet a target activated-parameter ratio $t \in [0, 1]$, we use a \textit{compression loss} similar to \citet{cai2024flextron}, defined over a batch of $N$ sequences of length $S$ as
\begin{equation}
\mathcal{L}_c = \max(0, c_{\mathsf{global}}  - t), 
\end{equation}
where 
\begin{equation}
c_{\mathsf{global}} = \sum_{b=1}^B \operatorname{softcap}_b(c_b)
\end{equation}
is a differentiable surrogate for the global activated-parameter ratio.
The blockwise soft cap approximates the cost of retaining a dense (i.e., non-factorized) block when its factorized implementation would be more expensive.
We define these terms below.

Let $y_{ijbm} \in \{0,1\}$ indicate whether token $j$ in sequence $i$ selects path $m$ in block $b$. The normalized average cost of block $b$ is
\begin{equation}
c_b = \frac{1}{NSP_0} \sum_{i=1}^{N} \sum_{j=1}^{S} \sum_{m=1}^{T} y_{ijbm} P_{bm},
\end{equation}
where
\begin{equation}
P_0 = \sum_{b=1}^{B} P_b^{\mathrm{dense}}
\end{equation}
is the total parameter count of the original weight matrices across all Transformer blocks (the embedding matrix and language-model output head are not counted, as they are not compressed).
Thus, $c_b$ is the contribution of block $b$ to the ratio of activated parameters relative to the original model. 
The normalized dense (non-factorized) cost of block $b$ is
\begin{equation}
C_b = \frac{P_b^{\mathrm{dense}}}{P_0}.
\end{equation}

The blockwise soft cap is defined as
\begin{equation}
\operatorname{softcap}_b(c_b)
=
c_b
-
\frac{C_b}{\kappa}
\operatorname{softplus}\!\left(
\frac{\kappa(c_b-C_b)}{C_b}
\right),
\end{equation}
where $\operatorname{softplus}(x)=\log(1+e^x)$ and $\kappa$ controls the slope. 
This function smoothly approximates $\min(c_b,C_b)$ and approaches $C_b$ as the factorized block's average cost increases above the dense cost.
Since $\sum_b C_b=1$, all block costs use the original model's Transformer weights as their common denominator.
A hard cap would limit the maximum cost to $C_b$, reflecting that inefficient blocks (\ie, where $c_b \geq C_b$) are retained in dense form at deployment.
However, it would also create a zero-gradient region, preventing the compression loss from providing an optimization signal.
The soft cap instead retains gradients above the dense cost, but is slightly lower than the hard-capped cost.
We therefore report actual deployment costs separately in \Cref{sec:mainAvgActParams,sec:avgActParams}.

During training, we freeze all model parameters except the routers and minimize the sum of the task and compression losses
\begin{equation}
\mathcal{L} = \mathcal{L}_{\mathrm{task}} + \lambda_s \mathcal{L}_{\mathrm{c}},
\end{equation}
where $\lambda_s$ is an adaptive compression-loss weight, updated at each training step $s$ according to the constraint violation as described in \Cref{sec:annealing}. The target-cost schedule and training hyperparameters are given in Appendix~\ref{app:implementationDetails}.

\section{Additional Experiments}\label{app:additionalResults}

\subsection{Additional Predictive Performance Results}\label{app:extraPerformance}
Similarly to the results presented in \Cref{sec:lm-performance}, we conduct an analysis of the downstream performance of \lrccblk{} compared to COALA, SVD-LLM, SVD-LLM, SVD-LLM+Swift, COALA+Swift, Swift-SVD (see \Cref{sec:lm-performance} for more details).
We present the complete results for \llamaSevenB{},  \llamaoneb{}  and \qwenThreeFourB{} respectively in   \Cref{tab:model-benchmarks-llama27b-256-seqlen2048-app}, \Cref{tab:model-benchmarks-llama321b-256-seqlen2048} and  \Cref{tab:model-benchmarks-qwen34b-256-seqlen2048}.
As can be observed from the two tables, \lrccblk{}-5 and \lrccblk{}-9 continue to surpass static methods in nearly all evaluations.
This demonstrates that models benefit from dynamically allocating different ranks depending on the tokens' needed accuracy, highlighting the strong assumptions static approaches do about models' computations.

\begin{figure}[h]
  \centering
  \includegraphics[width=0.6\linewidth]{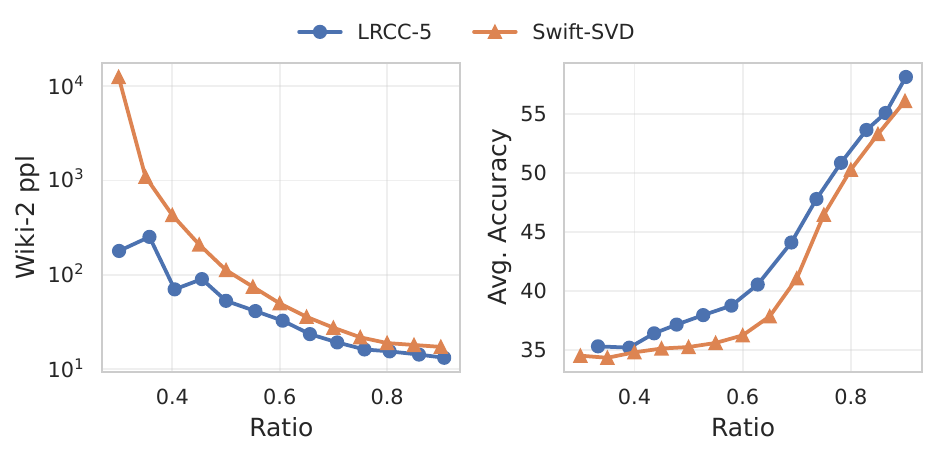}
  \caption{Pareto curves of \lrccblk{}-5 and \swiftsvd across various cost ratios applied to \qwenThreeFourB. The average accuracy is obtained by averaging the results on the commonsense reasoning downstream tasks.}
  \label{fig:paretoActiveRatioQwen}
\end{figure}

In \Cref{fig:paretoActiveRatioQwen} we compare \lrccblk{}-5 with \swiftsvd{} on \qwenThreeFourB{} across measured active-parameter ratios.
\lrccblk{} achieves lower perplexity at every ratio, with the largest gap under aggressive compression, where \swiftsvd{} degrades sharply.
On the downstream tasks, \lrccblk{} matches or exceeds \swiftsvd{} throughout, with the largest gains at intermediate ratios.
These trends mirror the Llama results in \Cref{fig:paretoActiveRatio}, suggesting that the benefits of \lrccblk{} are not specific to one model family.

\begin{table*}[h]
\centering
\begingroup
\normalfont
\footnotesize
\fontfamily{ptm}\selectfont
\setlength{\tabcolsep}{1.8pt}
\renewcommand{\arraystretch}{1.14}
\begin{adjustbox}{max width=\linewidth}
\begin{tabular}{l l<{\hspace{3pt}} *{8}{>{\scriptsize}r} r}
\toprule
& & \multicolumn{1}{c}{PPL $\downarrow$} & \multicolumn{8}{c}{Commonsense reasoning accuracy (\%) $\uparrow$} \\
\cmidrule(lr){3-3}\cmidrule(l){4-11}
Ratio & Method & \multicolumn{1}{c}{Wiki2} & \multicolumn{1}{c}{ARC-C} & \multicolumn{1}{c}{ARC-E} & \multicolumn{1}{c}{BoolQ} & \multicolumn{1}{c}{Hella} & \multicolumn{1}{c}{OBQA} & \multicolumn{1}{c}{PIQA} & \multicolumn{1}{c}{Wino} & \multicolumn{1}{c}{\textbf{Avg.}} \\
\midrule
\textcolor{black!60}{100\%} & \textcolor{black!60}{Base} & \textcolor{black!60}{5.49} & \benchref{44.80} & \benchref{73.95} & \benchref{79.42} & \benchref{76.08} & \benchref{44.00} & \benchref{78.62} & \benchref{69.30} & \benchref{66.60} \\
\midrule
\multirow{7}{*}[-1pt]{90\%} & COALA & 7.27 & \benchscore{35.5}{1.4} & \benchscore{57.5}{1.0} & \benchscore{49.3}{0.9} & \benchscore{62.8}{0.5} & \benchscore{38.4}{2.2} & \benchscore{70.6}{1.1} & \benchscore{65.2}{1.3} & \benchscore{54.2}{0.5} \\
 & SVD-LLM & 7.27 & \benchscore{36.2}{1.4} & \benchscore{57.4}{1.0} & \benchscore{50.1}{0.9} & \benchscore{62.8}{0.5} & \benchscore{39.0}{2.2} & \benchscore{70.8}{1.1} & \benchscore{65.3}{1.3} & \benchscore{54.5}{0.5} \\
 & SVD-LLM+Swift & 7.33 & \benchscore{35.5}{1.4} & \benchscore{58.3}{1.0} & \benchscore{55.0}{0.9} & \benchscore{62.8}{0.5} & \benchscore{38.4}{2.2} & \benchscore{71.1}{1.1} & \benchscore{62.9}{1.4} & \benchscore{54.8}{0.5} \\
 & COALA+Swift & 7.27 & \benchscore{35.1}{1.4} & \benchscore{58.4}{1.0} & \benchscore{57.4}{0.9} & \benchscore{62.8}{0.5} & \benchscore{38.4}{2.2} & \benchscore{70.9}{1.1} & \benchscore{63.1}{1.4} & \benchscore{55.1}{0.5} \\
 & Swift-SVD & 7.33 & \benchscore{35.6}{1.4} & \benchscore{58.7}{1.0} & \benchscore{55.4}{0.9} & \benchscore{62.8}{0.5} & \benchscore{39.6}{2.2} & \benchscore{70.7}{1.1} & \benchscore{62.9}{1.4} & \benchscore{55.1}{0.5} \\
\addlinespace[2pt]
\rowcolor{black!5}
\cellcolor{white} & \textbf{\lrccblk{}-5} & 6.17 & \benchscore{\textbf{41.8}}{1.4} & \benchscore{68.5}{1.0} & \benchscore{74.4}{0.8} & \benchscore{71.0}{0.5} & \benchscore{\textbf{41.4}}{2.2} & \benchscore{\textbf{74.8}}{1.0} & \benchscore{66.3}{1.3} & \benchscore{62.6}{0.5} \\
\rowcolor{black!5}
\cellcolor{white} & \textbf{\lrccblk{}-9} & \textbf{6.03} & \benchscore{40.4}{1.4} & \benchscore{\textbf{68.6}}{1.0} & \benchscore{\textbf{74.9}}{0.8} & \benchscore{\textbf{71.7}}{0.4} & \benchscore{40.8}{2.2} & \benchscore{\textbf{74.8}}{1.0} & \benchscore{\textbf{67.8}}{1.3} & \benchscore{\textbf{62.7}}{0.5} \\
\midrule
\multirow{7}{*}[-1pt]{70\%} & COALA & 10.75 & \benchscore{28.0}{1.3} & \benchscore{43.9}{1.0} & \benchscore{41.2}{0.9} & \benchscore{46.1}{0.5} & \benchscore{\textbf{35.0}}{2.1} & \benchscore{62.1}{1.1} & \benchscore{59.3}{1.4} & \benchscore{45.1}{0.5} \\
 & SVD-LLM & 10.75 & \benchscore{27.2}{1.3} & \benchscore{44.2}{1.0} & \benchscore{41.6}{0.9} & \benchscore{46.1}{0.5} & \benchscore{34.6}{2.1} & \benchscore{61.8}{1.1} & \benchscore{59.2}{1.4} & \benchscore{45.0}{0.5} \\
 & SVD-LLM+Swift & 10.58 & \benchscore{27.7}{1.3} & \benchscore{44.1}{1.0} & \benchscore{41.9}{0.9} & \benchscore{46.6}{0.5} & \benchscore{34.8}{2.1} & \benchscore{61.9}{1.1} & \benchscore{60.5}{1.4} & \benchscore{45.4}{0.5} \\
 & COALA+Swift & 10.58 & \benchscore{27.5}{1.3} & \benchscore{44.1}{1.0} & \benchscore{42.0}{0.9} & \benchscore{46.4}{0.5} & \benchscore{34.4}{2.1} & \benchscore{61.6}{1.1} & \benchscore{60.3}{1.4} & \benchscore{45.2}{0.5} \\
 & Swift-SVD & 10.58 & \benchscore{28.1}{1.3} & \benchscore{44.2}{1.0} & \benchscore{43.1}{0.9} & \benchscore{46.5}{0.5} & \benchscore{\textbf{35.0}}{2.1} & \benchscore{61.1}{1.1} & \benchscore{59.4}{1.4} & \benchscore{45.4}{0.5} \\
\addlinespace[2pt]
\rowcolor{black!5}
\cellcolor{white} & \textbf{\lrccblk{}-5} & 9.49 & \benchscore{31.4}{1.4} & \benchscore{46.3}{1.0} & \benchscore{\textbf{46.6}}{0.9} & \benchscore{56.0}{0.5} & \benchscore{34.6}{2.1} & \benchscore{\textbf{67.8}}{1.1} & \benchscore{\textbf{61.6}}{1.4} & \benchscore{\textbf{49.2}}{0.5} \\
\rowcolor{black!5}
\cellcolor{white} & \textbf{\lrccblk{}-9} & \textbf{9.05} & \benchscore{\textbf{31.9}}{1.4} & \benchscore{\textbf{46.4}}{1.0} & \benchscore{41.7}{0.9} & \benchscore{\textbf{56.5}}{0.5} & \benchscore{34.4}{2.1} & \benchscore{65.1}{1.1} & \benchscore{61.5}{1.4} & \benchscore{48.2}{0.5} \\
\midrule
\multirow{7}{*}[-1pt]{50\%} & COALA & 33.12 & \benchscore{22.7}{1.2} & \benchscore{30.9}{0.9} & \benchscore{\textbf{37.8}}{0.8} & \benchscore{31.0}{0.5} & \benchscore{26.6}{2.0} & \benchscore{53.2}{1.2} & \benchscore{53.9}{1.4} & \benchscore{36.6}{0.5} \\
 & SVD-LLM & 33.64 & \benchscore{23.0}{1.2} & \benchscore{30.9}{0.9} & \benchscore{\textbf{37.8}}{0.8} & \benchscore{31.1}{0.5} & \benchscore{27.0}{2.0} & \benchscore{53.0}{1.2} & \benchscore{54.5}{1.4} & \benchscore{36.8}{0.5} \\
 & SVD-LLM+Swift & \textbf{32.10} & \benchscore{23.0}{1.2} & \benchscore{30.8}{0.9} & \benchscore{\textbf{37.8}}{0.8} & \benchscore{31.3}{0.5} & \benchscore{26.6}{2.0} & \benchscore{53.1}{1.2} & \benchscore{54.7}{1.4} & \benchscore{36.8}{0.5} \\
 & COALA+Swift & \textbf{32.10} & \benchscore{23.0}{1.2} & \benchscore{30.5}{0.9} & \benchscore{\textbf{37.8}}{0.8} & \benchscore{31.4}{0.5} & \benchscore{26.6}{2.0} & \benchscore{52.8}{1.2} & \benchscore{54.6}{1.4} & \benchscore{36.7}{0.5} \\
 & Swift-SVD & 32.60 & \benchscore{23.1}{1.2} & \benchscore{30.9}{0.9} & \benchscore{\textbf{37.8}}{0.8} & \benchscore{31.4}{0.5} & \benchscore{27.0}{2.0} & \benchscore{52.4}{1.2} & \benchscore{\textbf{55.1}}{1.4} & \benchscore{36.8}{0.5} \\
\addlinespace[2pt]
\rowcolor{black!5}
\cellcolor{white} & \textbf{\lrccblk{}-5} & 41.86 & \benchscore{\textbf{24.8}}{1.3} & \benchscore{32.2}{1.0} & \benchscore{\textbf{37.8}}{0.8} & \benchscore{\textbf{32.7}}{0.5} & \benchscore{27.8}{2.0} & \benchscore{55.3}{1.2} & \benchscore{52.9}{1.4} & \benchscore{37.6}{0.5} \\
\rowcolor{black!5}
\cellcolor{white} & \textbf{\lrccblk{}-9} & 41.86 & \benchscore{22.8}{1.2} & \benchscore{\textbf{33.2}}{1.0} & \benchscore{\textbf{37.8}}{0.8} & \benchscore{32.1}{0.5} & \benchscore{\textbf{30.0}}{2.1} & \benchscore{\textbf{56.1}}{1.2} & \benchscore{53.0}{1.4} & \benchscore{\textbf{37.8}}{0.5} \\
\bottomrule
\end{tabular}%
\end{adjustbox}
\endgroup
\caption{Performance of Llama-2-7B compressed using 256 samples with sequence length 2048 on a language modeling dataset (measured using perplexity) and seven commonsense reasoning datasets (measured using accuracy).}
\label{tab:model-benchmarks-llama27b-256-seqlen2048-app}
\end{table*}

\begin{table*}[!t]
\centering
\begingroup
\normalfont
\footnotesize
\fontfamily{ptm}\selectfont
\setlength{\tabcolsep}{1.8pt}
\renewcommand{\arraystretch}{1.14}
\begin{adjustbox}{max width=\linewidth}
\begin{tabular}{l l<{\hspace{3pt}} *{8}{>{\scriptsize}r} r}
\toprule
& & \multicolumn{1}{c}{PPL $\downarrow$} & \multicolumn{8}{c}{Commonsense reasoning accuracy (\%) $\uparrow$} \\
\cmidrule(lr){3-3}\cmidrule(l){4-11}
Ratio & Method & \multicolumn{1}{c}{Wiki2} & \multicolumn{1}{c}{ARC-C} & \multicolumn{1}{c}{ARC-E} & \multicolumn{1}{c}{BoolQ} & \multicolumn{1}{c}{Hella} & \multicolumn{1}{c}{OBQA} & \multicolumn{1}{c}{PIQA} & \multicolumn{1}{c}{Wino} & \multicolumn{1}{c}{\textbf{Avg.}} \\
\midrule
\textcolor{black!60}{100\%} & \textcolor{black!60}{Base} & \textcolor{black!60}{9.79} & \benchref{36.69} & \benchref{61.95} & \benchref{63.61} & \benchref{64.23} & \benchref{37.00} & \benchref{74.65} & \benchref{61.01} & \benchref{57.02} \\
\midrule
\multirow{7}{*}[-1pt]{90\%} & COALA & 22.76 & \benchscore{25.9}{1.3} & \benchscore{38.4}{1.0} & \benchscore{48.3}{0.9} & \benchscore{35.0}{0.5} & \benchscore{27.0}{2.0} & \benchscore{56.4}{1.2} & \benchscore{52.6}{1.4} & \benchscore{40.5}{0.5} \\
 & SVD-LLM & 23.12 & \benchscore{25.6}{1.3} & \benchscore{38.7}{1.0} & \benchscore{47.6}{0.9} & \benchscore{34.8}{0.5} & \benchscore{27.6}{2.0} & \benchscore{56.4}{1.2} & \benchscore{52.3}{1.4} & \benchscore{40.4}{0.5} \\
 & SVD-LLM+Swift & 21.05 & \benchscore{26.7}{1.3} & \benchscore{41.3}{1.0} & \benchscore{54.3}{0.9} & \benchscore{38.9}{0.5} & \benchscore{28.2}{2.0} & \benchscore{59.4}{1.1} & \benchscore{55.5}{1.4} & \benchscore{43.5}{0.5} \\
 & COALA+Swift & 19.77 & \benchscore{26.5}{1.3} & \benchscore{44.4}{1.0} & \benchscore{\textbf{54.6}}{0.9} & \benchscore{42.1}{0.5} & \benchscore{31.2}{2.1} & \benchscore{62.2}{1.1} & \benchscore{56.2}{1.4} & \benchscore{45.3}{0.5} \\
 & Swift-SVD & 21.05 & \benchscore{25.7}{1.3} & \benchscore{38.8}{1.0} & \benchscore{51.0}{0.9} & \benchscore{36.7}{0.5} & \benchscore{27.6}{2.0} & \benchscore{57.7}{1.2} & \benchscore{52.6}{1.4} & \benchscore{41.4}{0.5} \\
\addlinespace[2pt]
\rowcolor{black!5}
\cellcolor{white} & \textbf{\lrccblk{}-5} & 11.81 & \benchscore{\textbf{32.6}}{1.4} & \benchscore{\textbf{54.2}}{1.0} & \benchscore{49.6}{0.9} & \benchscore{51.5}{0.5} & \benchscore{34.2}{2.1} & \benchscore{70.3}{1.1} & \benchscore{\textbf{56.7}}{1.4} & \benchscore{\textbf{49.9}}{0.5} \\
\rowcolor{black!5}
\cellcolor{white} & \textbf{\lrccblk{}-9} & \textbf{11.62} & \benchscore{31.3}{1.4} & \benchscore{53.1}{1.0} & \benchscore{46.3}{0.9} & \benchscore{\textbf{52.5}}{0.5} & \benchscore{\textbf{35.6}}{2.1} & \benchscore{\textbf{70.5}}{1.1} & \benchscore{54.7}{1.4} & \benchscore{49.1}{0.5} \\
\midrule
\multirow{7}{*}[-1pt]{70\%} & COALA & 209.30 & \benchscore{22.1}{1.2} & \benchscore{28.4}{0.9} & \benchscore{37.8}{0.8} & \benchscore{27.7}{0.4} & \benchscore{25.4}{1.9} & \benchscore{52.5}{1.2} & \benchscore{50.4}{1.4} & \benchscore{34.9}{0.5} \\
 & SVD-LLM & 222.79 & \benchscore{22.4}{1.2} & \benchscore{28.2}{0.9} & \benchscore{37.8}{0.8} & \benchscore{28.0}{0.4} & \benchscore{25.0}{1.9} & \benchscore{52.0}{1.2} & \benchscore{49.7}{1.4} & \benchscore{34.7}{0.5} \\
 & SVD-LLM+Swift & 135.13 & \benchscore{22.7}{1.2} & \benchscore{29.0}{0.9} & \benchscore{37.8}{0.8} & \benchscore{28.1}{0.4} & \benchscore{24.8}{1.9} & \benchscore{53.0}{1.2} & \benchscore{51.1}{1.4} & \benchscore{35.2}{0.5} \\
 & COALA+Swift & 130.97 & \benchscore{22.4}{1.2} & \benchscore{28.8}{0.9} & \benchscore{37.8}{0.8} & \benchscore{28.1}{0.4} & \benchscore{24.8}{1.9} & \benchscore{52.6}{1.2} & \benchscore{50.8}{1.4} & \benchscore{35.1}{0.5} \\
 & Swift-SVD & 157.98 & \benchscore{21.6}{1.2} & \benchscore{28.5}{0.9} & \benchscore{37.8}{0.8} & \benchscore{28.0}{0.4} & \benchscore{25.0}{1.9} & \benchscore{52.3}{1.2} & \benchscore{51.5}{1.4} & \benchscore{35.0}{0.5} \\
\addlinespace[2pt]
\rowcolor{black!5}
\cellcolor{white} & \textbf{\lrccblk{}-5} & 24.23 & \benchscore{22.7}{1.2} & \benchscore{37.8}{1.0} & \benchscore{38.6}{0.9} & \benchscore{\textbf{35.3}}{0.5} & \benchscore{27.4}{2.0} & \benchscore{58.2}{1.2} & \benchscore{50.3}{1.4} & \benchscore{38.6}{0.5} \\
\rowcolor{black!5}
\cellcolor{white} & \textbf{\lrccblk{}-9} & \textbf{23.48} & \benchscore{\textbf{24.9}}{1.3} & \benchscore{\textbf{38.7}}{1.0} & \benchscore{\textbf{42.0}}{0.9} & \benchscore{34.3}{0.5} & \benchscore{\textbf{28.2}}{2.0} & \benchscore{\textbf{59.6}}{1.1} & \benchscore{\textbf{51.9}}{1.4} & \benchscore{\textbf{40.0}}{0.5} \\
\midrule
\multirow{7}{*}[-1pt]{50\%} & COALA & 998.50 & \benchscore{22.6}{1.2} & \benchscore{25.6}{0.9} & \benchscore{37.8}{0.8} & \benchscore{27.0}{0.4} & \benchscore{25.6}{2.0} & \benchscore{49.6}{1.2} & \benchscore{50.4}{1.4} & \benchscore{34.1}{0.5} \\
 & SVD-LLM & 967.78 & \benchscore{22.9}{1.2} & \benchscore{25.8}{0.9} & \benchscore{37.8}{0.8} & \benchscore{27.0}{0.4} & \benchscore{25.2}{1.9} & \benchscore{49.9}{1.2} & \benchscore{50.3}{1.4} & \benchscore{34.1}{0.5} \\
 & SVD-LLM+Swift & 854.06 & \benchscore{\textbf{23.9}}{1.2} & \benchscore{26.1}{0.9} & \benchscore{37.8}{0.8} & \benchscore{26.7}{0.4} & \benchscore{24.4}{1.9} & \benchscore{50.1}{1.2} & \benchscore{49.3}{1.4} & \benchscore{34.1}{0.5} \\
 & COALA+Swift & 827.78 & \benchscore{23.4}{1.2} & \benchscore{26.2}{0.9} & \benchscore{37.8}{0.8} & \benchscore{26.7}{0.4} & \benchscore{24.6}{1.9} & \benchscore{50.1}{1.2} & \benchscore{49.5}{1.4} & \benchscore{34.1}{0.5} \\
 & Swift-SVD & 938.00 & \benchscore{23.3}{1.2} & \benchscore{26.3}{0.9} & \benchscore{37.8}{0.8} & \benchscore{26.5}{0.4} & \benchscore{24.8}{1.9} & \benchscore{51.1}{1.2} & \benchscore{50.1}{1.4} & \benchscore{34.3}{0.5} \\
\addlinespace[2pt]
\rowcolor{black!5}
\cellcolor{white} & \textbf{\lrccblk{}-5} & \textbf{79.44} & \benchscore{21.8}{1.2} & \benchscore{\textbf{33.9}}{1.0} & \benchscore{37.8}{0.8} & \benchscore{\textbf{29.2}}{0.5} & \benchscore{\textbf{26.8}}{2.0} & \benchscore{53.8}{1.2} & \benchscore{\textbf{50.6}}{1.4} & \benchscore{\textbf{36.3}}{0.5} \\
\rowcolor{black!5}
\cellcolor{white} & \textbf{\lrccblk{}-9} & 229.87 & \benchscore{21.2}{1.2} & \benchscore{33.7}{1.0} & \benchscore{\textbf{37.9}}{0.8} & \benchscore{27.8}{0.4} & \benchscore{25.6}{2.0} & \benchscore{\textbf{54.5}}{1.2} & \benchscore{49.9}{1.4} & \benchscore{35.8}{0.5} \\
\bottomrule
\end{tabular}%
\end{adjustbox}
\endgroup
\caption{Performance of Llama-3.2-1B compressed using 256 samples with sequence length 2048 on a language modeling dataset (measured using perplexity) and seven commonsense reasoning datasets (measured using accuracy).}
\label{tab:model-benchmarks-llama321b-256-seqlen2048}
\end{table*}

\begin{table*}[!t]
\centering
\begingroup
\normalfont
\footnotesize
\fontfamily{ptm}\selectfont
\setlength{\tabcolsep}{1.8pt}
\renewcommand{\arraystretch}{1.14}
\begin{adjustbox}{max width=\linewidth}
\begin{tabular}{l l<{\hspace{3pt}} *{8}{>{\scriptsize}r} r}
\toprule
& & \multicolumn{1}{c}{PPL $\downarrow$} & \multicolumn{8}{c}{Commonsense reasoning accuracy (\%) $\uparrow$} \\
\cmidrule(lr){3-3}\cmidrule(l){4-11}
Ratio & Method & \multicolumn{1}{c}{Wiki2} & \multicolumn{1}{c}{ARC-C} & \multicolumn{1}{c}{ARC-E} & \multicolumn{1}{c}{BoolQ} & \multicolumn{1}{c}{Hella} & \multicolumn{1}{c}{OBQA} & \multicolumn{1}{c}{PIQA} & \multicolumn{1}{c}{Wino} & \multicolumn{1}{c}{\textbf{Avg.}} \\
\midrule
& Swift-SVD & 17.18 & \benchscore{38.1}{1.4} & \benchscore{56.4}{1.0} & \benchscore{76.2}{0.7} & \benchscore{58.1}{0.5} & \benchscore{34.8}{2.1} & \benchscore{68.4}{1.1} & \benchscore{60.7}{1.4} & \benchscore{56.1}{0.5} \\
\addlinespace[2pt]
\rowcolor{black!5}
\cellcolor{white} & \textbf{\lrccblk{}-5} & 13.17 & \benchscore{41.1}{1.4} & \benchscore{63.5}{1.0} & \benchscore{\textbf{77.3}}{0.7} & \benchscore{58.0}{0.5} & \benchscore{\textbf{35.2}}{2.1} & \benchscore{70.3}{1.1} & \benchscore{\textbf{61.3}}{1.4} & \benchscore{\textbf{58.1}}{0.5} \\
\rowcolor{black!5}
\cellcolor{white}\multirow{-3}{*}{90\%} & \textbf{\lrccblk{}-9} & \textbf{12.77} & \benchscore{\textbf{42.7}}{1.4} & \benchscore{\textbf{65.5}}{1.0} & \benchscore{68.8}{0.8} & \benchscore{\textbf{59.0}}{0.5} & \benchscore{\textbf{35.2}}{2.1} & \benchscore{\textbf{71.7}}{1.1} & \benchscore{59.9}{1.4} & \benchscore{57.5}{0.5} \\
\midrule
& Swift-SVD & 27.45 & \benchscore{25.0}{1.3} & \benchscore{37.9}{1.0} & \benchscore{48.5}{0.9} & \benchscore{37.1}{0.5} & \benchscore{29.2}{2.0} & \benchscore{57.9}{1.2} & \benchscore{51.9}{1.4} & \benchscore{41.1}{0.5} \\
\addlinespace[2pt]
\rowcolor{black!5}
\cellcolor{white} & \textbf{\lrccblk{}-5} & 19.17 & \benchscore{27.9}{1.3} & \benchscore{46.5}{1.0} & \benchscore{\textbf{60.8}}{0.9} & \benchscore{\textbf{44.9}}{0.5} & \benchscore{30.8}{2.1} & \benchscore{62.7}{1.1} & \benchscore{\textbf{53.0}}{1.4} & \benchscore{\textbf{46.7}}{0.5} \\
\rowcolor{black!5}
\cellcolor{white}\multirow{-3}{*}{70\%} & \textbf{\lrccblk{}-9} & \textbf{18.00} & \benchscore{\textbf{29.1}}{1.3} & \benchscore{\textbf{48.1}}{1.0} & \benchscore{45.9}{0.9} & \benchscore{42.2}{0.5} & \benchscore{\textbf{34.4}}{2.1} & \benchscore{\textbf{65.2}}{1.1} & \benchscore{52.7}{1.4} & \benchscore{45.4}{0.5} \\
\midrule
 & Swift-SVD & 112.03 & \benchscore{\textbf{23.9}}{1.2} & \benchscore{28.2}{0.9} & \benchscore{\textbf{38.2}}{0.8} & \benchscore{29.4}{0.5} & \benchscore{24.0}{1.9} & \benchscore{52.9}{1.2} & \benchscore{50.3}{1.4} & \benchscore{35.3}{0.5} \\
\addlinespace[2pt]
\rowcolor{black!5}
\cellcolor{white} & \textbf{\lrccblk{}-5} & 52.92 & \benchscore{23.6}{1.2} & \benchscore{\textbf{36.4}}{1.0} & \benchscore{37.8}{0.8} & \benchscore{32.4}{0.5} & \benchscore{\textbf{27.4}}{2.0} & \benchscore{\textbf{55.5}}{1.2} & \benchscore{\textbf{51.2}}{1.4} & \benchscore{\textbf{37.8}}{0.5} \\
\rowcolor{black!5}
\cellcolor{white}\multirow{-3}{*}{50\%} & \textbf{\lrccblk{}-9} & \textbf{45.26} & \benchscore{21.9}{1.2} & \benchscore{32.9}{1.0} & \benchscore{37.8}{0.8} & \benchscore{\textbf{32.5}}{0.5} & \benchscore{25.0}{1.9} & \benchscore{54.0}{1.2} & \benchscore{50.9}{1.4} & \benchscore{36.4}{0.5} \\
\bottomrule
\end{tabular}%
\end{adjustbox}
\endgroup
\caption{Performance of Qwen3-4B compressed using 256 samples with sequence length 2048 on a language modeling dataset (measured using perplexity) and seven commonsense reasoning datasets (measured using accuracy).}
\label{tab:model-benchmarks-qwen34b-256-seqlen2048}
\end{table*}

\subsection{Computational Performance Results}\label{sec:addResCompute}
This section provides the measurement protocol and per-task latency results for \Cref{sec:mainComputePerformance}.
We measured the latency of \lrccblk{} compared with COALA+Swift, a representative low-rank compression approach, across several target costs and downstream tasks.
We teacher-forced the ground-truth task tokens through the model to ensure identical inputs for both deployments and measured latency as the time required to decode each complete sequence divided by its number of decoded tokens. 
For multiple-choice tasks, we timed each candidate continuation and treated all continuations associated with the same document as one sample cluster.
All measurements used a batch size of 1.

The latency estimates are based on four randomly selected 2,048-token chunks from \wikitext{}, 16 documents from WinoGrande, eight from OpenBookQA, and four from each remaining task: ARC-Challenge, ARC-Easy, BoolQ, HellaSwag, and PIQA. 
These sample counts were selected based on pilot validation runs, with additional samples allocated to datasets exhibiting greater document-level variability. 
Paired document-cluster bootstrap intervals for the final workloads had relative half-widths of at most \(2.14\%\). 
All candidate continuations from a sampled document were kept together during resampling to avoid treating correlated choices as independent observations.

Both methods were compiled using TorchDynamo and TorchInductor with identical settings.
CUDA graphs were disabled because we found their support limited for conditional methods and to ensure a fair comparison. 
We performed three 64-token warm-up runs before each measurement to exclude compilation and initialization from the timed region (compilation and prefill times are explored in Appendix~\ref{sec:appendixNumberPaths}).
Experiments were conducted on an NVIDIA GB10 Grace Blackwell with 128\,GB of unified memory.

We repeated each complete experiment five times to estimate run-to-run variability. 
Each table entry reports the mean latency in milliseconds per token across the five executions, and the subscripted \(\pm\) term gives the half-width of a 95\% Student-\(t\) confidence interval computed from the five execution-level measurements (\(n=5\), \(df=4\)). 
\Cref{tab:decode-latencies-llama27b-lrccblk5} and \Cref{tab:decode-latencies-llama321b-lrccblk5} show the results.

At the same nominal target cost, \lrccblk{} is slower on \llamaSevenB{}, while the ordering varies with the target cost on \llamaoneb{}.
\lrccblk{} introduces overhead from evaluating the router and dispatching the selected computational path, but its executed cost also depends on the routing decisions.
As discussed in \Cref{sec:results}, at matched latencies, \lrccblk{} produces competitive predictive performance and outperforms the baseline in several settings. 
Furthermore, our implementation does not use specialized fused CUDA kernels, leaving opportunities for further speedups. 
For example, fusing the router with layer normalization could reduce the number of times the residual stream is read from memory.

\begin{table*}[ht]
\centering
\footnotesize
\setlength{\tabcolsep}{2pt}
\begin{adjustbox}{max width=\linewidth}
\begin{tabular}{llcccccccc}
\toprule
Target & Method & Wiki2 & ARC-C & ARC-E & BoolQ & Hella & OBQA & PIQA & Wino \\
\midrule
90\% & COALA+ & \err{66.98}{0.08} & \err{67.01}{0.09} & \err{67.01}{0.09} & \err{66.99}{0.08} & \err{67.00}{0.09} & \err{67.02}{0.10} & \err{67.01}{0.09} & \err{67.01}{0.09} \\
 & \lrccblk{}-5 & \err{70.83}{0.13} & \err{70.96}{0.12} & \err{70.98}{0.12} & \err{71.32}{0.13} & \err{71.51}{0.13} & \err{70.54}{0.11} & \err{71.37}{0.12} & \err{70.36}{0.11} \\
\addlinespace[2pt]
80\% & COALA+ & \err{60.29}{0.10} & \err{60.32}{0.11} & \err{60.33}{0.11} & \err{60.31}{0.10} & \err{60.32}{0.10} & \err{60.33}{0.11} & \err{60.35}{0.15} & \err{60.33}{0.11} \\
 & \lrccblk{}-5 & \err{64.73}{0.10} & \err{64.86}{0.11} & \err{65.10}{0.11} & \err{65.00}{0.10} & \err{65.34}{0.11} & \err{65.21}{0.11} & \err{65.71}{0.11} & \err{65.01}{0.11} \\
\addlinespace[2pt]
70\% & COALA+ & \err{54.31}{0.07} & \err{54.34}{0.07} & \err{54.37}{0.06} & \err{54.35}{0.06} & \err{54.34}{0.06} & \err{54.38}{0.07} & \err{54.34}{0.06} & \err{54.36}{0.07} \\
 & \lrccblk{}-5 & \err{57.96}{0.04} & \err{59.19}{0.07} & \err{59.53}{0.07} & \err{58.54}{0.06} & \err{59.25}{0.06} & \err{59.54}{0.08} & \err{60.30}{0.07} & \err{59.15}{0.07} \\
\addlinespace[2pt]
60\% & COALA+ & \err{48.09}{0.07} & \err{48.13}{0.09} & \err{48.14}{0.10} & \err{48.12}{0.08} & \err{48.12}{0.08} & \err{48.12}{0.08} & \err{48.12}{0.08} & \err{48.12}{0.08} \\
 & \lrccblk{}-5 & \err{51.05}{0.05} & \err{53.06}{0.06} & \err{53.14}{0.06} & \err{51.87}{0.05} & \err{52.56}{0.05} & \err{53.12}{0.07} & \err{53.37}{0.06} & \err{52.34}{0.06} \\
\addlinespace[2pt]
50\% & COALA+ & \err{41.89}{0.04} & \err{41.91}{0.05} & \err{41.91}{0.05} & \err{41.92}{0.05} & \err{41.92}{0.05} & \err{41.89}{0.05} & \err{41.90}{0.04} & \err{41.89}{0.04} \\
 & \lrccblk{}-5 & \err{44.39}{0.06} & \err{45.31}{0.11} & \err{45.28}{0.09} & \err{44.74}{0.08} & \err{45.00}{0.09} & \err{44.64}{0.11} & \err{45.43}{0.10} & \err{44.85}{0.11} \\
\addlinespace[2pt]
40\% & COALA+ & \err{35.76}{0.08} & \err{35.79}{0.07} & \err{35.78}{0.07} & \err{35.80}{0.07} & \err{35.79}{0.07} & \err{35.76}{0.07} & \err{35.78}{0.06} & \err{35.77}{0.07} \\
 & \lrccblk{}-5 & \err{37.73}{0.08} & \err{38.63}{0.10} & \err{38.88}{0.10} & \err{38.30}{0.09} & \err{38.41}{0.10} & \err{38.12}{0.12} & \err{39.02}{0.10} & \err{38.12}{0.12} \\
\addlinespace[2pt]
30\% & COALA+ & \err{29.35}{0.08} & \err{29.41}{0.02} & \err{29.41}{0.02} & \err{29.42}{0.02} & \err{29.42}{0.02} & \err{29.41}{0.03} & \err{29.41}{0.02} & \err{29.41}{0.03} \\
 & \lrccblk{}-5 & \err{30.82}{0.08} & \err{32.08}{0.09} & \err{32.43}{0.10} & \err{31.58}{0.08} & \err{31.81}{0.09} & \err{31.39}{0.11} & \err{32.69}{0.17} & \err{31.72}{0.12} \\
\bottomrule
\end{tabular}
\end{adjustbox}
\caption{Decode latency (ms/token) for \lrccblk{}-5 and COALA+Swift (shown as COALA+) at matched target costs for \llamaSevenB{} at batch size 1. 
Each value corresponds to the mean latency across five benchmark executions.
We also show the 95\% Student-$t$ confidence interval half-width ($n=5$, $df=4$).
Both methods use identical tokenized inputs in each execution.}
\label{tab:decode-latencies-llama27b-lrccblk5}
\end{table*}

\begin{table*}[ht]
\centering
\footnotesize
\setlength{\tabcolsep}{2pt}
\begin{adjustbox}{max width=\linewidth}
\begin{tabular}{llcccccccc}
\toprule
Target & Method & Wiki2 & ARC-C & ARC-E & BoolQ & Hella & OBQA & PIQA & Wino \\
\midrule
90\% & COALA+ & \err{15.34}{0.07} & \err{15.37}{0.07} & \err{15.35}{0.07} & \err{15.36}{0.07} & \err{15.36}{0.07} & \err{15.35}{0.07} & \err{15.35}{0.07} & \err{15.35}{0.07} \\
 & \lrccblk{}-5 & \err{15.77}{0.09} & \err{15.98}{0.07} & \err{15.94}{0.07} & \err{15.90}{0.07} & \err{15.97}{0.07} & \err{16.02}{0.06} & \err{16.03}{0.07} & \err{15.96}{0.06} \\
\addlinespace[2pt]
80\% & COALA+ & \err{16.37}{0.07} & \err{16.34}{0.06} & \err{16.33}{0.06} & \err{16.38}{0.06} & \err{16.36}{0.06} & \err{16.28}{0.06} & \err{16.32}{0.06} & \err{16.29}{0.06} \\
 & \lrccblk{}-5 & \err{14.90}{0.04} & \err{15.21}{0.05} & \err{15.27}{0.05} & \err{15.09}{0.05} & \err{15.18}{0.05} & \err{15.47}{0.04} & \err{15.27}{0.04} & \err{15.40}{0.04} \\
\addlinespace[2pt]
70\% & COALA+ & \err{15.15}{0.07} & \err{15.12}{0.07} & \err{15.11}{0.07} & \err{15.15}{0.07} & \err{15.13}{0.07} & \err{15.06}{0.07} & \err{15.10}{0.07} & \err{15.06}{0.07} \\
 & \lrccblk{}-5 & \err{14.18}{0.11} & \err{14.61}{0.11} & \err{14.62}{0.11} & \err{14.42}{0.11} & \err{14.53}{0.11} & \err{15.06}{0.12} & \err{14.82}{0.11} & \err{14.74}{0.12} \\
\addlinespace[2pt]
60\% & COALA+ & \err{14.13}{0.08} & \err{14.10}{0.08} & \err{14.10}{0.08} & \err{14.14}{0.08} & \err{14.13}{0.08} & \err{14.05}{0.08} & \err{14.09}{0.08} & \err{14.05}{0.08} \\
 & \lrccblk{}-5 & \err{13.22}{0.05} & \err{13.84}{0.07} & \err{13.96}{0.07} & \err{13.57}{0.07} & \err{13.64}{0.07} & \err{14.40}{0.07} & \err{14.00}{0.07} & \err{13.92}{0.07} \\
\addlinespace[2pt]
50\% & COALA+ & \err{13.41}{0.09} & \err{13.36}{0.08} & \err{13.35}{0.08} & \err{13.41}{0.08} & \err{13.39}{0.08} & \err{13.28}{0.08} & \err{13.34}{0.08} & \err{13.28}{0.08} \\
 & \lrccblk{}-5 & \err{12.68}{0.04} & \err{13.55}{0.05} & \err{13.51}{0.05} & \err{13.03}{0.04} & \err{13.17}{0.05} & \err{14.03}{0.07} & \err{13.76}{0.06} & \err{13.38}{0.07} \\
\addlinespace[2pt]
40\% & COALA+ & \err{12.41}{0.08} & \err{12.38}{0.08} & \err{12.37}{0.08} & \err{12.41}{0.08} & \err{12.40}{0.08} & \err{12.32}{0.08} & \err{12.36}{0.08} & \err{12.33}{0.08} \\
 & \lrccblk{}-5 & \err{13.56}{0.05} & \err{14.12}{0.05} & \err{14.23}{0.05} & \err{13.78}{0.07} & \err{13.74}{0.06} & \err{14.40}{0.07} & \err{14.31}{0.07} & \err{13.89}{0.06} \\
\addlinespace[2pt]
30\% & COALA+ & \err{11.59}{0.04} & \err{11.55}{0.04} & \err{11.54}{0.04} & \err{11.59}{0.04} & \err{11.57}{0.04} & \err{11.48}{0.04} & \err{11.53}{0.04} & \err{11.49}{0.04} \\
 & \lrccblk{}-5 & \err{12.55}{0.05} & \err{12.95}{0.06} & \err{13.10}{0.05} & \err{12.73}{0.06} & \err{12.69}{0.07} & \err{13.17}{0.06} & \err{13.07}{0.05} & \err{12.85}{0.07} \\
\bottomrule
\end{tabular}
\end{adjustbox}
\caption{Decode latency (ms/token) for \lrccblk{}-5 and COALA+Swift (shown as COALA+) at matched target costs for \llamaoneb{} at batch size 1. 
Each value corresponds to the mean latency across five benchmark executions.
We also show the 95\% Student-$t$ confidence interval half-width ($n=5$, $df=4$).
Both methods use identical tokenized inputs in each execution.}
\label{tab:decode-latencies-llama321b-lrccblk5}
\end{table*}

\subsection{Additional Results on Target vs. Actual Deployment Cost}\label{sec:avgActParams}
\begin{figure*}[!t]
  \centering
  \includegraphics[width=0.98\textwidth]{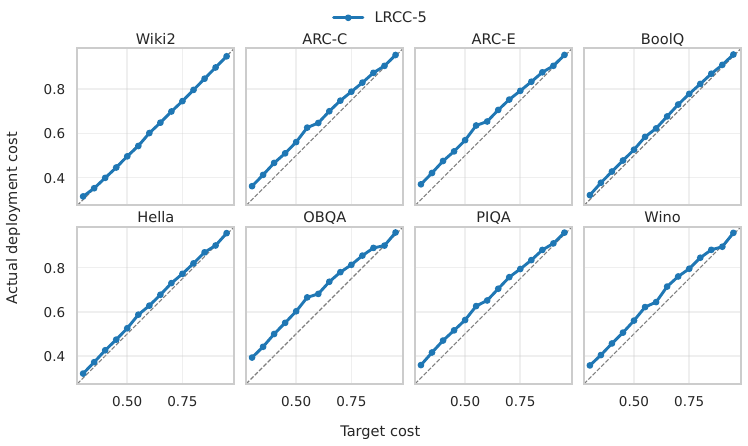}
  \caption{Target versus actual deployment cost for \llamaoneb{} with \lrccblk{}-5. Each panel shows one evaluation task; the dashed line indicates exact agreement between the target and actual deployment cost.}
  \label{fig:deploy-cost-target-vs-actual-llama321b}
\end{figure*}

\begin{figure*}[!t]
  \centering
  \includegraphics[width=0.98\textwidth]{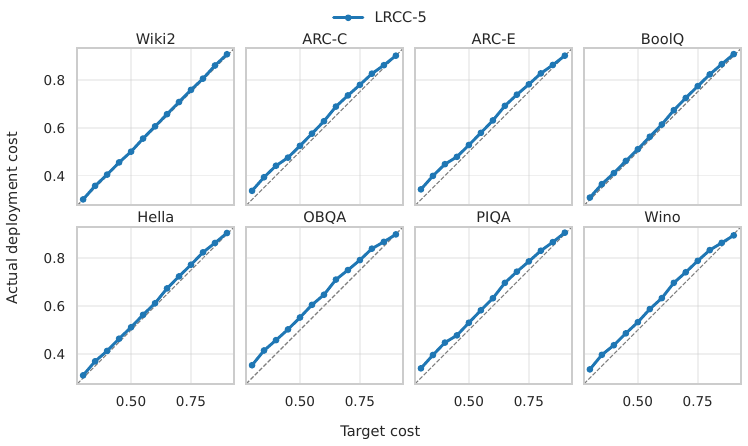}
  \caption{Target versus actual deployment cost for Qwen3-4B with \lrccblk{}-5. Each panel shows one evaluation task; the dashed line indicates exact agreement between the target and actual deployment cost.}
  \label{fig:deploy-cost-target-vs-actual-qwen34b}
\end{figure*}

Due to the input-dependence of the \lrcc method, the average ratio of activated parameters is not strictly the same as the specified in the target cost.
\Cref{tab:deploy-cost-models-lrcc5-9-256-seqlen2048} shows the actual ratio of activated parameters in the different datasets, corresponding to the same checkpoints whose results were reported in Appendix~\ref{app:additionalResults} and for \llamaSevenB{} in \Cref{tab:model-benchmarks-llama27b-256-seqlen2048}.
\begin{table*}[!t]
\centering
\small
\setlength{\tabcolsep}{2pt}
\begin{tabular}{lllccccccccc}
\toprule
Ratio & Method & Model & Wiki2 & ARC-C & ARC-E & BoolQ & Hella & OBQA & PIQA & Wino & Avg. \\
\midrule
\multirow{6}{*}{90\%}
& \lrccblk{}-5 & Llama-2-7B & 0.90 & 0.86 & 0.86 & 0.86 & 0.86 & 0.90 & 0.87 & 0.90 & 0.87 \\
& \lrccblk{}-9 & Llama-2-7B & 0.90 & 0.87 & 0.87 & 0.87 & 0.87 & 0.87 & 0.88 & 0.87 & 0.87 \\
\softcmidrule
& \lrccblk{}-5 & Llama-3.2-1B & 0.90 & 0.87 & 0.88 & 0.87 & 0.87 & 0.90 & 0.88 & 0.89 & 0.88 \\
& \lrccblk{}-9 & Llama-3.2-1B & 0.90 & 0.86 & 0.86 & 0.86 & 0.86 & 0.88 & 0.87 & 0.87 & 0.87 \\
\softcmidrule
& \lrccblk{}-5 & Qwen3-4B & 0.91 & 0.90 & 0.90 & 0.91 & 0.90 & 0.90 & 0.91 & 0.89 & 0.90 \\
& \lrccblk{}-9 & Qwen3-4B & 0.91 & 0.91 & 0.91 & 0.91 & 0.91 & 0.91 & 0.91 & 0.91 & 0.91 \\
\midrule
\multirow{6}{*}{70\%}
& \lrccblk{}-5 & Llama-2-7B & 0.70 & 0.68 & 0.69 & 0.67 & 0.68 & 0.69 & 0.69 & 0.68 & 0.68 \\
& \lrccblk{}-9 & Llama-2-7B & 0.70 & 0.70 & 0.70 & 0.67 & 0.68 & 0.70 & 0.66 & 0.69 & 0.68 \\
\softcmidrule
& \lrccblk{}-5 & Llama-3.2-1B & 0.70 & 0.65 & 0.65 & 0.68 & 0.68 & 0.68 & 0.65 & 0.64 & 0.66 \\
& \lrccblk{}-9 & Llama-3.2-1B & 0.70 & 0.70 & 0.66 & 0.68 & 0.68 & 0.69 & 0.65 & 0.70 & 0.68 \\
\softcmidrule
& \lrccblk{}-5 & Qwen3-4B & 0.71 & 0.69 & 0.69 & 0.72 & 0.72 & 0.71 & 0.70 & 0.70 & 0.70 \\
& \lrccblk{}-9 & Qwen3-4B & 0.71 & 0.69 & 0.69 & 0.73 & 0.67 & 0.71 & 0.70 & 0.69 & 0.70 \\
\midrule
\multirow{6}{*}{50\%}
& \lrccblk{}-5 & Llama-2-7B & 0.45 & 0.47 & 0.47 & 0.46 & 0.46 & 0.48 & 0.47 & 0.47 & 0.47 \\
& \lrccblk{}-9 & Llama-2-7B & 0.45 & 0.47 & 0.47 & 0.46 & 0.46 & 0.48 & 0.47 & 0.47 & 0.47 \\
\softcmidrule
& \lrccblk{}-5 & Llama-3.2-1B & 0.50 & 0.47 & 0.47 & 0.48 & 0.47 & 0.50 & 0.47 & 0.46 & 0.47 \\
& \lrccblk{}-9 & Llama-3.2-1B & 0.45 & 0.49 & 0.49 & 0.46 & 0.46 & 0.45 & 0.49 & 0.49 & 0.48 \\
\softcmidrule
& \lrccblk{}-5 & Qwen3-4B & 0.50 & 0.52 & 0.53 & 0.51 & 0.51 & 0.50 & 0.48 & 0.49 & 0.51 \\
& \lrccblk{}-9 & Qwen3-4B & 0.50 & 0.49 & 0.49 & 0.51 & 0.51 & 0.50 & 0.49 & 0.48 & 0.50 \\
\bottomrule
\end{tabular}
\caption{Actual deployment cost (activated-parameter ratio) for the \lrccblk{}-5 and \lrccblk{}-9 configurations used during evaluation.}
\label{tab:deploy-cost-models-lrcc5-9-256-seqlen2048}
\end{table*}
Additionally, \Cref{fig:deploy-cost-target-vs-actual-llama321b} and \Cref{fig:deploy-cost-target-vs-actual-qwen34b} show the actual deployment costs versus the target cost used during training for \llamaoneb{} and \qwenThreeFourB{}, similarly to \Cref{sec:mainAvgActParams}.

\subsection{Ablation on the Number of Paths}\label{sec:appendixNumberPaths}
\begin{figure}[h]
  \centering
  \begin{subfigure}[t]{0.48\textwidth}
    \centering
    \includegraphics[width=\linewidth]{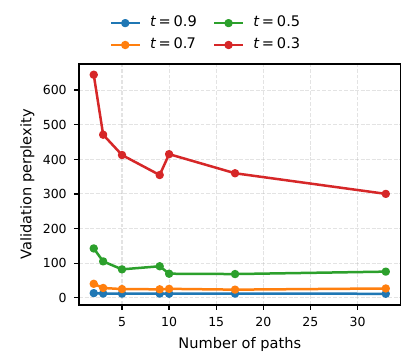}
    \caption{}
  \end{subfigure}
  \hfill
  \begin{subfigure}[t]{0.48\textwidth}
    \centering
    \includegraphics[width=\linewidth]{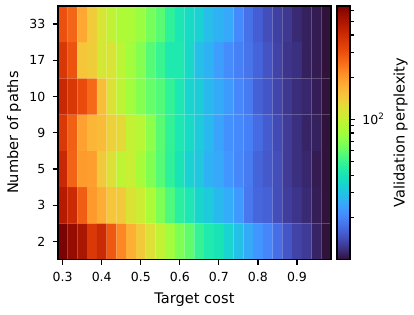}
    \caption{}
  \end{subfigure}
  \caption{
    Ablation on the number of paths for \llamaoneb. (a) shows \wikitext{} validation perplexity as a function of the number of paths for selected target costs, while (b) shows the results for all target costs from $0.3$ to $0.975$ in steps of $0.025$.
  }
  \label{fig:numberPathsLlama32App}
\end{figure}

As described in \Cref{sec:mainNumberPaths}, we trained models with \(P\in\{2,3,5,9,10,17,33\}\) paths per block to understand the influence of this hyperparameter.
Each path corresponds to a rank assignment with a target cost uniformly spaced between \(0.1\) and \(1.0\).
At each refinement, we retain the previous paths and add paths with target costs midway between them.
For example, with \(P=2\), we use paths constructed by the rank-allocation algorithm with target costs of \(0.1\) and \(1.0\).
For \(P=3\), we add a path with target cost \(0.55\).
The results for \llamaSevenB{} are shown in \Cref{fig:numberPathsLlama2}.
\Cref{fig:numberPathsLlama32App} shows the perplexities for \llamaoneb{}.
Perplexity increases sharply with only two paths, but increasing the number of choices beyond five provides comparatively little improvement.
This agrees with the results for \llamaSevenB{} in \Cref{sec:mainNumberPaths}.

\begin{figure}[!h]
  \centering
  \begin{subfigure}[t]{0.98\textwidth}
    \centering
    \includegraphics[width=\linewidth]{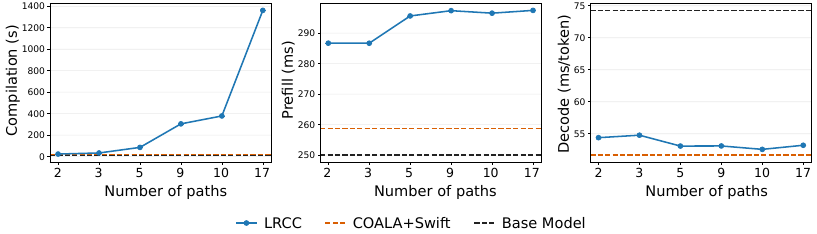}
    \caption{\llamaSevenB}
  \end{subfigure}
  \vfil
  \begin{subfigure}[t]{0.98\textwidth}
    \centering
    \includegraphics[width=\linewidth]{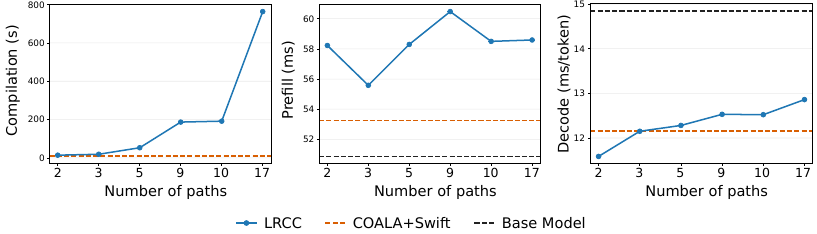}
    \caption{\llamaoneb}
  \end{subfigure}
  \caption{
    Effect of the number of paths on compilation, prefill, and decoding times. Increasing the number of paths leads to substantially higher compilation times and slightly higher decoding times.
  }
  \label{fig:numberPathsTimings}
\end{figure}

We also evaluate the effect of the number of paths on compilation, prefill with 1,024 tokens, and decoding for models with a target cost of \(0.7\).
For this ablation, we decode 64 tokens after prefilling with samples from \wikitext{}, rather than using the complete task sequences timed in \Cref{sec:addResCompute}.
\Cref{fig:numberPathsTimings} shows that compilation time is the metric most affected by the number of paths.
Compiling the 33-path models was infeasible because graph capture exceeded TorchDynamo's recursion depth.
Compilation also became substantially slower as the number of paths increased.
Increasing the number of paths has little effect on the decoding and prefill latencies of \llamaSevenB{} but increases the decoding latency of \llamaoneb{}.
Overall, the 5-path models provide a good trade-off between predictive and computational performance.
Finally, we note that the prefill implementation of \lrccblk{} computes the first factor multiplication at full rank, masks the intermediate representation according to the selected path, and then applies the second factor, thus performing computation for ranks that are subsequently masked.
The prefill implementation therefore does not yet exploit all of the computation savings available from routing; we leave its optimization to future work.
Other optimizations could improve the prefill performance of both static low-rank and LRCC models relative to the dense base model, for example by fusing operations that share the same inputs \citep{zhu2026different}.

\subsection{Router Choices}\label{app:routerChoices}
\begin{figure}[h]
  \centering
  \begin{subfigure}[t]{0.470\textwidth}
    \centering
    \includegraphics[width=\linewidth]{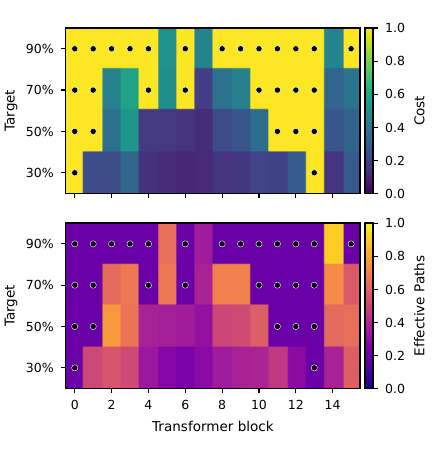}
    \caption{}
  \end{subfigure}
  \begin{subfigure}[t]{0.508\textwidth}
    \centering
    \includegraphics[width=\linewidth]{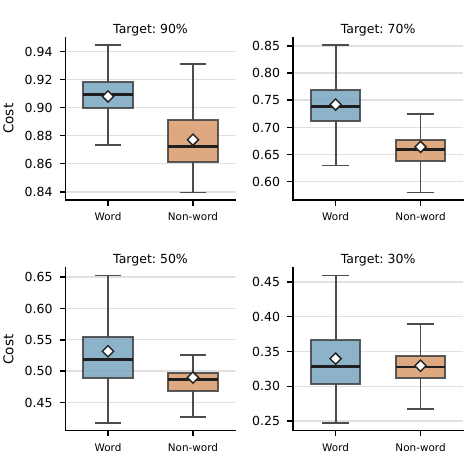}
    \caption{}
  \end{subfigure}
  \caption{Router choices for \llamaoneb{} on \wikitext{}. (a) Average per-block cost (top) and effective fraction of paths (bottom) for each target. The dots indicate blocks that are deployed without factorization. (b) Distributions of the costs assigned to word and non-word tokens for each target.}
  \label{fig:routerChoicesLlama32}
\end{figure}

We repeated the same experiments described in \Cref{sec:routerChoices} using \llamaoneb{}.
The results are shown in \Cref{fig:routerChoicesLlama32}.
As in \llamaSevenB{}, a backbone of blocks is retained in dense form, while other Transformer blocks use multiple paths across tokens.
\Cref{fig:routerChoicesLlama32}(b) also suggests that \NonWord{} tokens are assigned less compute on average.

\section{Implementation Details}\label{app:implementationDetails}
All training was conducted on A100 GPUs with 64GB of VRAM.
We trained linear block-level routers on 256 \wikitext{} sequences of length 2,048 with BF16 parameters using Adam with a constant learning rate of $10^{-3}$ and effective batch size of 4.
Training annealed over 28 target costs, from $0.975$ to $0.300$ in steps of \(-0.025\) with 12 epochs per target by default and up to 30 epochs when no valid checkpoint was found.
A training epoch took, on average, 2 minutes 24 seconds for \llamaSevenB{} and 28 seconds for \llamaoneb{}.
Both models required approximately 350 epochs to produce all 28 checkpoints, corresponding to approximately 14 hours for \llamaSevenB{} and 2.3 hours for \llamaoneb{}, or about 30 and 5 minutes per checkpoint, respectively.

For the compression loss, we set the soft-cap slope to $\kappa = 20$.
We initialized the compression-loss weight at $\lambda_0 = 1$.
We selected the dual learning rate \(\alpha\) for \(\lambda\) and the Gumbel-Softmax temperature schedule using a model-specific grid of four configurations. 
For Llama-3.2-1B, we searched dual learning rates \(\alpha \in \{0.4,0.8\}\) and temperature schedules of a fixed $\tau = 0.5$ and \(2.0\!\rightarrow\!0.5\).
For \llamaSevenB{}, we searched \(\alpha \in \{1.2,1.6\}\) and temperature schedules of a fixed $\tau = 2.0$ and \(2.0\!\rightarrow\!0.5\).
Among the four runs, we selected the configuration with the lowest \wikitext{} validation loss at the largest number of the 28 target costs.
We repeated this search independently for each path count in \(\{2,3,5,9,10,17,33\}\).

Using the authors' source code, we reimplemented COALA, Swift-SVD, and SVD-LLM in our framework to ensure consistent experimental settings and evaluation.
We omit SVD-LLM's optional fine-tuning stage so that no method fine-tunes the pretrained weights after compression; \lrccblk{} additionally trains its routers. An analogous weight-refinement stage could be applied to \lrccblk{} and the baselines.
For COALA, we searched accumulation values $\{2, 4\}$ and regularization values $\mu \in \{0, 1\}$, at each target cost in $\{0.3, 0.5, 0.7, 0.9\}$.
We found that an accumulation value of 2 and $\mu=0$ gave the best \wikitext{} validation performance, although differences were small. We used these values for all COALA decompositions, including the base factorization of the Llama models in \lrccblk{}.
For Swift-SVD, we used their recommended layer importance weight of $0.6$ in \llamaSevenB{}.
For \llamaoneb{}, we observed that the best results were obtained with a layer importance weight of $1.0$.
We used the recommended defaults for the remaining hyperparameters.
SVD-LLM required no hyperparameter tuning.
Finally, COALA+Swift and SVD-LLM+Swift combine the respective decompositions with the rank mappings produced by Swift-SVD.
We refer to the original papers for definitions of the baseline-specific hyperparameters.

\subsection{Efficient Computation}\label{app:efficientComputation}
The optimal factors require only the leading right singular vectors
of the layer outputs
$\mathbf{Y}=\mathbf{X}\mathbf{W}$, and different methods compute this
subspace in different ways.
Swift-SVD \citep{qi2026swiftsvd}, for example, avoids storing the full
output matrix by incrementally accumulating its Gram matrix over
calibration batches,
\begin{equation}
    \mathbf{G}_{Y}
    =
    \mathbf{Y}^{\top}\mathbf{Y}
    =
    \sum_j
    \mathbf{Y}_j^{\top}\mathbf{Y}_j,
    \qquad
    \mathbf{Y}_j=\mathbf{X}_j\mathbf{W}.
\end{equation}
Since
\begin{equation}
    \mathbf{G}_{Y}
    =
    \mathbf{V}\mathbf{\Sigma}^2\mathbf{V}^{\top},
\end{equation}
a single eigendecomposition of the accumulated matrix directly
provides $\mathbf{V}$ and the squared singular values.

FlexRank \citep{zaccone2026flexrank} uses a closely related strategy,
but instead incrementally accumulates the second moment of the
\emph{input} activations,
\begin{equation}
    \mathbf{G}_{X}
    =
    \mathbf{X}^{\top}\mathbf{X}
    =
    \sum_j
    \mathbf{X}_j^{\top}\mathbf{X}_j.
\end{equation}
Writing $\mathbf{S}=\mathbf{G}_{X}^{1/2}$, the right singular vectors
of $\mathbf{S}\mathbf{W}$ are the same as those of the complete
output matrix because
\begin{equation}
    (\mathbf{S}\mathbf{W})^{\top}
    (\mathbf{S}\mathbf{W})
    =
    \mathbf{W}^{\top}\mathbf{X}^{\top}
    \mathbf{X}\mathbf{W}
    =
    \mathbf{Y}^{\top}\mathbf{Y}.
\end{equation}
This makes the memory required for calibration independent of the
number of calibration tokens.

COALA \citep{parkina2026coala} instead proposes avoiding the explicit
formation of a Gram matrix. In our matrix convention, let
\begin{equation}
    \mathbf{X}=\mathbf{Q}\mathbf{R}
\end{equation}
be a thin QR decomposition of the calibration activations, computed
with tall-skinny QR (TSQR) when $\mathbf{X}$ is too large to
materialize at once. Since $\mathbf{Q}^{\top}\mathbf{Q}=\mathbf{I}$,
\begin{equation}
    \left\|
        \mathbf{X}(\mathbf{W}-\widetilde{\mathbf{W}})
    \right\|_F
    =
    \left\|
        \mathbf{R}(\mathbf{W}-\widetilde{\mathbf{W}})
    \right\|_F.
\end{equation}
Furthermore,
\begin{equation}
    (\mathbf{R}\mathbf{W})^{\top}
    (\mathbf{R}\mathbf{W})
    =
    \mathbf{W}^{\top}\mathbf{R}^{\top}\mathbf{R}\mathbf{W}
    =
    \mathbf{W}^{\top}\mathbf{X}^{\top}\mathbf{X}\mathbf{W}
    =
    \mathbf{Y}^{\top}\mathbf{Y}.
\end{equation}
Therefore, an SVD of the substantially smaller matrix
$\mathbf{R}\mathbf{W}$ yields the same right singular subspace
$\mathbf{V}$ in exact arithmetic. Unlike approaches based on the
normal equations, this procedure does not explicitly form
$\mathbf{X}^{\top}\mathbf{X}$ or require its inversion. This is
numerically advantageous for ill-conditioned calibration
activations, since forming a Gram matrix squares the condition number
and can amplify floating-point errors.

In our experiments, we used COALA's QR strategy to compute the decompositions for the Llama models. For \qwenThreeFourB{}, the COALA decomposition did not converge, so we accumulated the output Gram matrix over calibration batches following \citet{qi2026swiftsvd}.

\end{document}